%% file: acl_latex.tex
\documentclass[11pt]{article}

\usepackage[final]{acl}

\usepackage{times}
\usepackage{latexsym}

\usepackage[T1]{fontenc}

\usepackage[utf8]{inputenc}
\usepackage{amsfonts}

\usepackage{microtype}

\usepackage{inconsolata}

\usepackage{graphicx}
\usepackage{subcaption}
\usepackage{amsmath}
\usepackage{booktabs}
\usepackage{multirow}

\title{LLM Judges Can Encode More About Summary Quality Than Their Scores Report}

\author{Rajarshi Haldar \quad Ikhyun Cho \quad Sewoong Lee \quad Julia Hockenmaier \\
  Siebel School of Computing and Data Science \\
  University of Illinois Urbana-Champaign, Urbana, IL, USA \\
  \texttt{\{rhaldar2,ihcho2,samuel27,juliahmr\}@illinois.edu}}

\usepackage{tcolorbox}
\tcbuselibrary{skins,breakable}
\usepackage{listings}

\definecolor{accContrib}{HTML}{0B7A57}
\definecolor{accRecord}{HTML}{A05F00}
\definecolor{accPrompt}{HTML}{8E2F7E}
\definecolor{lstink}{RGB}{35,35,35}

\tcbset{calloutbase/.style 2 args={
  enhanced, colframe=#1, colback=#1!5!white,
  colbacktitle=#1, coltitle=white,
  fonttitle=\bfseries\footnotesize,
  title={#2},
  boxrule=0pt, leftrule=2.5pt, arc=2.5pt, outer arc=2.5pt,
  left=6pt, right=5pt, top=5pt, bottom=5pt, boxsep=0pt,
  toptitle=2.5pt, bottomtitle=2.5pt, lefttitle=6pt,
  before skip=6pt, after skip=6pt}}

\newtcolorbox{findings}{calloutbase={accContrib}{Key findings},
  breakable, use color stack, title after break={},
  top=3pt, bottom=3pt, before skip=4pt, after skip=4pt}
\newcommand{\frun}[1]{\textbf{\textcolor{accContrib}{#1}}}

\newcommand{\lstcaptionbar}[2]{%
  \noindent\makebox[\linewidth][l]{%
    \colorbox{#1}{\textcolor{white}{\bfseries\footnotesize\strut~#2~}}}%
  \par\nobreak\vspace{-3pt}}
\lstdefinestyle{promptbase}{
  basicstyle=\ttfamily\footnotesize\color{lstink},
  columns=fullflexible, keepspaces=true, showstringspaces=false,
  breaklines=false, escapeinside={(*@}{@*)},
  frame=l, framerule=2.5pt,
  framesep=7pt, xleftmargin=10pt, xrightmargin=2pt,
  aboveskip=2pt, belowskip=2pt}
\lstdefinestyle{recordlisting}{style=promptbase,
  backgroundcolor=\color{accRecord!5!white}, rulecolor=\color{accRecord}}
\lstdefinestyle{promptlisting}{style=promptbase,
  backgroundcolor=\color{accPrompt!5!white}, rulecolor=\color{accPrompt}}
\newcommand{\lsthead}[1]{\textcolor{accPrompt}{\bfseries #1}}
\newcommand{\lstrechead}[1]{\textcolor{accRecord}{\bfseries #1}}
\newcommand{\lstslotname}[1]{\textcolor{accRecord}{\bfseries #1}}

\begin{document}
\maketitle
\begin{abstract}
LLM-as-a-judge systems are assessed mainly by how well their scores agree with human ratings.
Instead of treating the judge as a black box, we ask what it represents internally when it applies an evaluation criterion.
We use sparse autoencoders to examine the internal states of Gemma~3 and Qwen~3.5 while they score SummEval summaries, and isolate features that separate high- from low-scoring summaries.
We find these features to be few, concentrated in later layers, and criterion-specific.
We then fit probes on expert ratings, using a few dozen such features per probe.
These probes are competitive with probes fit on the full hidden state, and usually agree with expert ratings better than the judge's reported scores.
We conclude that the reported score can be a poorer indicator of summary quality than the judge's activations.
\end{abstract}

\section{Introduction}
\label{sec:intro}

Large language models (LLMs) now routinely score the output of other models in place of human raters~\citep{zheng2023judging,gu2025surveyllmasajudge}.
These LLM judges are then typically assessed by how far their ratings agree with human ratings~\citep{bavaresco-etal-2025-llms,validating_llm_judge}.
This verification typically uses only the judge's final output score and treats the judge as a black box.

Prior audits show that judges are swayed by the order in which two candidates are compared~\citep{wang-etal-2024-large-language-models-fair}, by style~\citep{wu-aji-2025-style}, and by their own authorship~\citep{panickssery2024llm}.
Judges have also been observed to give the same input different scores on different runs~\citep{stureborg2024inconsistent,haldar-hockenmaier-2025-rating}.
A low agreement score indicates that something went wrong, but it does not tell us what exactly it is that went wrong.
It cannot say whether the judge failed to form an assessment or failed to report it.
Therefore, we ask a question that cannot be answered by just looking at the reported score.
\emph{What does a judge represent internally when it applies an evaluation criterion, and does that representation apply the criterion the same way a human evaluator does?}
To answer this question, we have to go beyond the final score assigned by the judge and analyze its internal activations.
To analyze a model's internals, we use sparse autoencoders (SAEs), whose features are likelier to correspond to something more meaningful than the judge's hidden state activations~\citep{elhage2022toy,sae_find_highly}.
Such features have been previously mapped to concrete concepts whose presence is easy to check. For example, one feature fires on mentions of the Golden Gate Bridge~\citep{templeton2026}.
On the other hand, an evaluation criterion is more abstract and its presence is not as simple to check, since a criterion like coherence is evaluated on a passage as a whole.

We prompt Gemma~3~\citep{gemma3technicalreport} and Qwen~3.5~\citep{qwen35technicalreport} to score the 1,600 summaries of the SummEval benchmark on its four quality criteria.
We then analyze these judges with the Gemma Scope~2~\citep{gemmascope2} and Qwen-Scope~\citep{deng2026qwenscopeturningsparsefeatures} sparse autoencoders.
We isolate the features whose firing tracks expert-annotated quality, then fit probes on them.
In our case, a probe is a simple ordinal classifier that can test what a given representation actually encodes.
Our code is publicly available.\footnote{\url{https://github.com/rajarshihaldar/probing-summary-judges}}

\begin{findings}
\frun{Very few features track expert-annotated quality overall, with more features being present in the later layers examined} (42-87 of the 132-152 features that ever fire for Gemma~3 and 44-75 of the 182-213 for Qwen~3.5).
\newline
\frun{Judges can distinguish between different criteria and prioritize identifying flaws in the input.} We find that each criterion has its own selective feature set, and more features identify bad summaries than good ones.
\newline
\frun{Activations contain more information than the final output score.} Supervised probes over a few dozen features beat the judge's own reported score in most cases. For the criterion fluency, the agreement improves from negative to positive.
\end{findings}

\section{Background}
\label{sec:background}

To recover what an LLM judge represents, we need access to that judge's internal activations, a benchmark with separate labels for multiple criteria, and a tool for inspecting activations.

\paragraph{LLM judges and their decoded scores.}
An LLM-as-a-judge is prompted with a rubric and a piece of text, and returns a rating on a fixed scale~\citep{zheng2023judging,gu2025surveyllmasajudge}.
We call that rating the \emph{decoded score}.
Those ratings are ordinal rather than continuous, so agreement is measured with Krippendorff's $\alpha$ rather than a correlation coefficient~\citep{Krippendorff2011ComputingKA,artstein-poesio-2008-survey}.
An $\alpha$ of 1 is perfect agreement and 0 is completely random, and values in between mean agreement above chance.
A negative $\alpha$ means the two sources disagree systematically rather than at random.

\paragraph{The SummEval benchmark.}
SummEval is a summarization benchmark of 1,600 news summaries, produced by 16 systems over 100 CNN/DailyMail articles~\citep{fabbri-etal-2021-summeval}, and Appendix~\ref{app:summeval_details} documents it in full.
Every summary carries multiple human ratings on four criteria adapted from the DUC summarization guidelines~\citep{dang-2006-duc}, \textit{coherence}, \textit{consistency}, \textit{fluency} and \textit{relevance}.
\emph{Coherence} measures whether the sentences in the summary are logically connected, internally consistent and easy to follow, and \emph{consistency} measures whether any statement contradicts the source.
\emph{Fluency} measures whether individual sentences are well formed, and \emph{relevance} measures whether the summary keeps the important content.
Two pools of raters evaluated every summary on all four criteria on a 1-5 scale: three \emph{experts} recruited for a background in summarization research, and five crowd workers, the \emph{turkers}, recruited through Amazon Mechanical Turk.
We take the median of the expert ratings as the ground truth.

\paragraph{Activations and sparse autoencoders.}
As an LLM judge processes its prompt, every layer writes a vector into the \emph{residual stream}, the running representation the model carries forward from one layer to the next.
This means that whatever a judge understands about the summary and the rubric has to be present in its activations before it assigns a rating.
Inspecting it directly is difficult, because individual neurons are polysemantic, meaning they activate for multiple unrelated concepts.
This is because these models have to keep track of more distinct properties than they have dimensions, a phenomenon known as \emph{superposition}~\citep{elhage2022toy}.
A \emph{sparse autoencoder} (SAE) addresses it by re-expressing the same vector in a much larger basis where only a few coordinates are non-zero at once~\citep{sae_find_highly,shu-etal-2025-survey}.
Each coordinate of that basis is a \emph{feature}, and the number of coordinates is the SAE's \emph{width}, or equivalently its dictionary size.
Pretrained SAE suites released alongside several open models let us decompose a judge's activations without training an SAE ourselves~\citep{lieberum-etal-2024-gemma,gemmascope2,deng2026qwenscopeturningsparsefeatures}.

\paragraph{Active and selective features.}
We label a feature as \emph{active} on a set of items as long as it has a non-zero activation on at least one of them.
Active features are already a small share of a dictionary.
Most of them respond to generic properties of the prompt such as syntax or token identity, firing whichever summary the judge is scoring.
A feature is \emph{selective} if it does something stronger, firing disproportionately on the summaries experts rated well, or disproportionately on the ones they rated badly.
The active set counts how much of the dictionary the judge uses at all.
Meanwhile, a selective feature is one that tracks expert-annotated quality.

\paragraph{Probes.}
A \emph{probe} is a simple supervised model fit on a frozen representation in order to test what that representation encodes~\citep{alain2017understanding,belinkov-2022-probing}.
If a probe can correctly predict a property, in our case the expert-assigned rating, that means the information to do so was present in the activations.
However, a sufficiently expressive probe can learn the task itself instead of recovering it from the representation~\citep{hewitt-liang-2019-designing}. To avoid this, we use simple ordinal regression probes.

\section{Methodology}
\label{sec:methodology}

Our experiments answer four research questions.
Do features that track summary quality exist at all, and where can they be found in the network?
Does an LLM judge represent quality as the presence of merit or as the absence of flaws?
Does it distinguish between the four criteria, or apply one notion of quality to all of them?
And do a judge's residual activations contain enough information to predict expert judgment better than its output score?
Our experimental setup has two stages.
We first identify selective features across all configurations to choose a configuration, taking the number of features and compute and storage efficiency into account.
Only then do we fit probes to test the predictive power of selective features under that configuration.

\subsection{Scoring and Feature Extraction}
\label{sec:extraction}

We evaluate each summary once per criterion and save both the decoded score and the residual activations.
For each input we prompt the judge with the criterion definition, a five-level rubric, and the summary, following the form-filling format of G-Eval~\citep{liu-etal-2023-g}.
The source article is provided for every criterion except fluency, which SummEval defines over the summary alone.
Appendix~\ref{app:judgments} gives the full prompt for every criterion.

To obtain the decoded score, we use the judge's distribution of scores.
At the first position after the prompt we take the logits for the five tokens \texttt{"1"} to \texttt{"5"} and renormalize them with a softmax over those five alone.
We then round the expectation $\sum_{s=1}^{5} s \cdot P(s)$ to the nearest integer.
This is G-Eval's procedure, which is deterministic and avoids the arbitrary tie-breaking of a sampled integer~\citep{liu-etal-2023-g}.
That rounded expectation is the decoded score.

To get the activations, we record the residual stream at the last token position of the prompt, the position immediately before the judge commits to a rating, at every layer.
Each recorded vector then goes through a pretrained SAE for that model and layer, giving one sparse feature vector per summary, criterion, layer and width.
We use Gemma Scope~2 for Gemma~3~\citep{gemmascope2} and Qwen-Scope for Qwen~3.5~\citep{deng2026qwenscopeturningsparsefeatures}.
We extract SAE features at layers 16, 31, 40 and 53, the four depths at which Gemma Scope~2 releases SAEs at all of its widths.
We used the same four layers for Qwen~3.5 to keep the comparison consistent, although Qwen-Scope offers SAEs for all layers.
For Gemma~3 we use widths 16k, 65k and 262k, while for Qwen-Scope we use its single available width of 80k.
We report our probing results on layer 53, at 16k for Gemma~3 and 80k for Qwen~3.5.
Gemma~3 has 62 layers and Qwen~3.5 has 64, so layer 53 is near the output in both models.
In Appendix~\ref{app:all_selected} we show the feature counts for every available layer and width.
We limited our experiments and our claims to these four layers due to time and resource constraints.

\subsection{Splitting the Data}
\label{sec:splits}

Because we select features using the expert labels, the split has to come before any measurement.
For each criterion we partition the 1,600 summaries into 1,280 training and 320 test examples~\citep{pedregosa2011scikit}.
We stratify the split on the expert rating, so that each of the five levels has the same distribution of scores in both sets.
Adding a further constraint on article or system identity would make the number of items at each rating level too sparse.
We therefore do not group the split, so a source article or one of the 16 systems may appear on both sides.
A probe could predict the score from the article or system identity rather than summary quality.
In Appendix~\ref{app:surface_baseline} we consider this possibility, and in the Limitations section we discuss other possible implications of this.
We will use the training set to identify selective SAE features and also train our probes on those features.

Within the training set we also sort summaries into two quality groups.
A summary is \emph{high-scoring} for a criterion if its median expert rating on that criterion is 4 or 5, and \emph{low-scoring} if it is 1, 2 or 3.
The grouping is done per criterion, since a summary rated highly coherent need not be equally fluent.

\subsection{Identifying Selective Features}
\label{sec:selectivity_method}

A feature is selective when it fires disproportionately on either the high-scoring or low-scoring group.
But first we need to ask two questions: what counts as disproportionate, and how much evidence we need to consider it disproportionate.

For the first, we record each feature as on or off for every training summary, where on means a non-zero activation.
Let $f_{\text{high}}$ and $f_{\text{low}}$ be the numbers of high- and low-scoring summaries a feature fires on, respectively.
We want to find $p_{\text{high}} = f_{\text{high}} / (f_{\text{high}} + f_{\text{low}})$, the share of that feature's firings landing on high-scoring summaries.
Let $N_{\text{high}}$ and $N_{\text{low}}$ be the numbers of high- and low-scoring training summaries.
A feature that is irrelevant to summary quality would have $p_{\text{high}}$ equal to the base rate $\pi_0 = N_{\text{high}} / (N_{\text{high}} + N_{\text{low}})$, the share of training summaries that are high-scoring.
Comparing the two directly is misleading, because the variance of a proportion depends on where it falls in $[0,1]$.
Therefore, we measure the gap with Cohen's $h$, an arcsine-stabilized effect size:

\begin{equation}
h = 2\left(\arcsin\sqrt{p_{\text{high}}} - \arcsin\sqrt{\pi_0}\right),
\end{equation}

and keep features with $|h| > 0.2$, which Cohen calls a small effect~\citep{cohen1988statistical}.
We set the bar low on purpose, since the question is whether quality-relevant structure is present at all.

We also require a two-sided Fisher's exact test on the two-by-two table of firing against quality group to reject independence at the $5\%$ level~\citep{fisher1922interpretation}.
We apply this test separately to each feature without correction for multiple comparisons.
Our split fixes both margins, and a feature fires at most once per summary.
This makes the null hypergeometric, which makes Fisher's test the right one (Appendix~\ref{app:selectivity_test}).
A feature counts as \emph{selective} for a criterion when both conditions hold. We evaluate the conditions separately for every criterion and judge.

\subsection{Fitting Probes}
\label{sec:probe_method}

Now that we have identified selective features, we need to know if the information contained in them can be used to predict the expert rating.
To assess this, we fit one ordinal regression probe per criterion and feature set, using \texttt{mord}'s \texttt{LogisticAT}~\citep{pedregosa2017consistency}.
An ordinal probe respects the ordering of the 1-5 scale rather than treating the five levels as unrelated classes.
Probes are scored against the median expert rating on the 320 test summaries.

Probes take the continuous SAE activations as input, not the binarized firing used for selection.
Every feature is centered to zero mean and scaled to unit variance.
We fix the regularization strength (\texttt{mord}'s $\alpha$) at 10.0 for each feature set. Larger values penalize more.
Changing this $\alpha$ from 0.01 to 1000 did not affect the ranking across different feature sets, so the results do not depend on this choice.
Uncertainty comes from 10{,}000 bootstrap resamples of the test summaries, with the probe fit held fixed.
Every set is scored on the same resample, so each yields a paired difference.

\section{Experiments and Results}
\label{sec:results}

Before getting into the performance of our trained probes, we need to know how well the decoded score from the LLM judges agrees with expert ratings as a baseline.

\subsection{How Well Do the Judges Score?}
\label{sec:judge_agreement}

We begin by measuring each judge from its output, that is, its decoded score. To do this, we compare its decoded score against the median expert rating on the 320 test summaries.

\begin{table}[ht]
  \footnotesize
\centering
\begin{tabular}{lrr}
\toprule
Criterion & Gemma-3 & Qwen-3.5 \\
\midrule
Coherence   & 0.500  & 0.590  \\
Consistency & 0.639  & 0.532  \\
Fluency     & $-0.419$ & $-0.106$ \\
Relevance   & 0.480  & 0.521  \\
\bottomrule
\end{tabular}
\caption{Krippendorff's $\alpha$ (ordinal) between each judge's decoded score and the median expert rating, by criterion, on the 320 test summaries of Section~\ref{sec:splits}.}
\label{tab:judge_agreement}
\end{table}

For both judges, we observe agreement values between $0.480$ and $0.639$ on three of the four criteria (Table~\ref{tab:judge_agreement}), falling short of the matched expert ceiling on metrics like coherence and consistency (Appendix~\ref{app:expert_ceiling}).
Fluency, on the other hand, shows negative agreement for both judges.
This means that the judges disagree with the experts more often than chance assignment would.
We see that the crowd annotations also fall below zero on fluency, at $-0.39$ over the full dataset, so this failure can also occur with untrained human judges (Appendix~\ref{app:source_agreement}).
Our own rubric may contribute, since it asks for no formatting problems and SummEval's summaries are uniformly lowercased.
This is the anomaly Section~\ref{sec:probing_results} explains, and it is the reason we analyze the activations at all.

\subsection{Do Selective Features Exist, and Where?}
\label{sec:feature_activation}

Before asking what a judge's features encode, we have to establish whether any of them even track summary quality.

Across the four layers examined, selective features are scarce in the early layers of the LLM judge and are more concentrated in the later layers (Appendix~\ref{app:all_selected}), which is why we set aside layers 16 and 31.
Gemma~3's feature counts for coherence are 0, 3, 64 and 74 across the four layer positions.
For Qwen~3.5, we observe more features at the lowest layer and a dip at layer 40 for every criterion.
The trend is clear for one judge and uneven for the other.
We also see that SAE width choice barely matters.
At layer 53 the narrowest Gemma Scope~2 dictionary returns at least as many selective features as the widest on three of the four criteria, even though narrower dictionaries reconstruct less faithfully in general~\citep{gao2025scaling}.

\begin{table}[ht]
  \footnotesize
\centering
\begin{tabular}{lrr}
\toprule
Criterion & Gemma-3 & Qwen-3.5 \\
\midrule
Coherence   & 74 & 57 \\
Consistency & 87 & 75 \\
Fluency     & 42 & 60 \\
Relevance   & 79 & 44 \\
\bottomrule
\end{tabular}
\caption{Selective features per criterion at layer 53, using Gemma Scope~2 (width 16k) for Gemma~3 and Qwen-Scope (width 80k) for Qwen~3.5.}
\label{tab:selective_counts}
\end{table}

Selecting features as Section~\ref{sec:selectivity_method} describes returns a small set for every judge and criterion (Table~\ref{tab:selective_counts}).
Those sets lie within a set that is itself small.
Gemma~3 activates 132 to 152 features per criterion and Qwen~3.5 activates 182 to 213, which is at most $0.9\%$ of Gemma Scope~2's 16,384 features and $0.3\%$ of Qwen-Scope's 81,920.
Taken as a union across all 1{,}600 summaries, that set is about three times the per-prompt sparsity for Gemma~3 and twice that for Qwen~3.5.
Most of it responds to the summary rather than to the prompt template (Appendix~\ref{app:sae-setup}).
Across the two judges, between $24\%$ and $57\%$ of active features are also selective.
This shows that the number of features actually required to discern between high-quality and low-quality summaries is in the dozens.
However, whether these features are qualitatively meaningful to us is a different question.
We looked at Neuronpedia's explanations for Gemma~3's selective features and they refer to numbers and code rather than to any summary-evaluating criteria (Appendix~\ref{app:neuronpedia}).

\subsection{Merit, or the Absence of Flaws?}
\label{sec:exp11}

Knowing that selective features exist does not tell us anything about what they are actually detecting.
A feature can fire on the summaries experts rated well or on the ones they rated badly.
Which of those is more common can give us insight into how an LLM judge comes up with a rating.

\begin{figure}[ht]
\centering
\includegraphics[width=\linewidth]{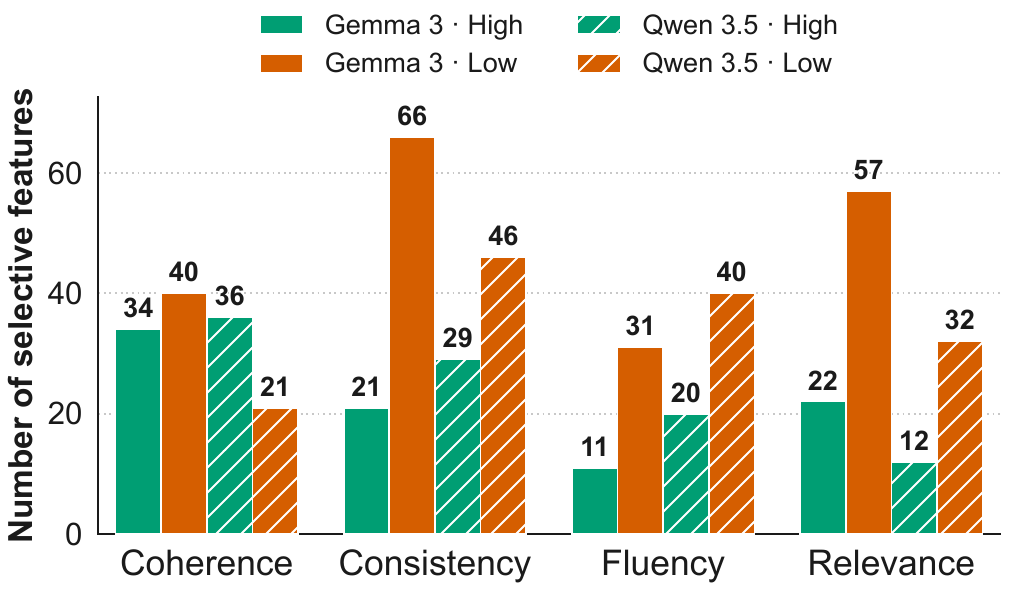}
\caption{Features selective for high- versus low-scoring summaries, where high and low are the experts' ratings and not the judge's, by criterion and judge.
In all cases but one, more features identify low-scoring summaries.}
\label{fig:high_low}
\end{figure}

Features that identify low-scoring summaries outnumber those that identify high-scoring ones in seven of the eight judge-criterion pairs (Figure~\ref{fig:high_low}).
The margin is often wide.
Gemma~3 has 66 features identifying inconsistent summaries against 21 identifying consistent ones.
Coherence under Qwen~3.5 is the single exception, with 36 features identifying coherent summaries against 21 identifying incoherent ones.
More of what a judge represents about a criterion tracks what goes wrong in a summary than what goes right.

\begin{table*}[t]
\footnotesize
\centering
\renewcommand{\arraystretch}{0.75}
\begin{tabular}{ll rrr rrr}
\toprule
& & \multicolumn{3}{c}{Gemma-3-27b-it} & \multicolumn{3}{c}{Qwen-3.5-27b} \\
\cmidrule(lr){3-5} \cmidrule(lr){6-8}
Criterion & Feature Set & N Features & $\alpha$ & sd$(\Delta)$ & N Features & $\alpha$ & sd$(\Delta)$ \\
\midrule
\multirow{6}{*}{Coherence}
 & Judge (Decoded) & -- & $0.500$ & $0.034$ & -- & $0.590$ & $0.034$ \\
 & Hidden State Activations & 5376 & $0.679$ & $0.031$ & 5120 & $0.679$ & $0.032$ \\
 & All Active & 146 & $0.645$ & $0.021$ & 191 & $0.670$ & $0.022$ \\
 & Active-Not-Selective & 72 & $0.622$ & $0.028$ & 134 & $0.639$ & $0.028$ \\
 & Selective-Union & 140 & $0.670$ & $0.020$ & 142 & $0.668$ & $0.022$ \\
 & Selective & 74 & $\mathbf{0.681}$ & -- & 57 & $\mathbf{0.683}$ & -- \\
\midrule
\multirow{6}{*}{Consistency}
 & Judge (Decoded) & -- & $\mathbf{0.639}$ & $0.032$ & -- & $0.532$ & $0.032$ \\
 & Hidden State Activations & 5376 & $0.555$ & $0.030$ & 5120 & $0.615$ & $0.031$ \\
 & All Active & 152 & $0.523$ & $0.021$ & 213 & $0.615$ & $0.024$ \\
 & Active-Not-Selective & 65 & $0.515$ & $0.028$ & 138 & $0.594$ & $0.029$ \\
 & Selective-Union & 140 & $0.537$ & $0.021$ & 142 & $0.617$ & $0.020$ \\
 & Selective & 87 & $0.557$ & -- & 75 & $\mathbf{0.694}$ & -- \\
\midrule
\multirow{6}{*}{Fluency}
 & Judge (Decoded) & -- & $-0.419$ & $0.040$ & -- & $-0.106$ & $0.036$ \\
 & Hidden State Activations & 5376 & $\mathbf{0.581}$ & $0.038$ & 5120 & $0.587$ & $0.033$ \\
 & All Active & 132 & $0.551$ & $0.023$ & 208 & $0.574$ & $0.023$ \\
 & Active-Not-Selective & 90 & $0.424$ & $0.026$ & 148 & $0.560$ & $0.030$ \\
 & Selective-Union & 140 & $0.553$ & $0.022$ & 142 & $0.575$ & $0.021$ \\
 & Selective & 42 & $0.558$ & -- & 60 & $\mathbf{0.600}$ & -- \\
\midrule
\multirow{6}{*}{Relevance}
 & Judge (Decoded) & -- & $0.480$ & $0.034$ & -- & $0.521$ & $0.038$ \\
 & Hidden State Activations & 5376 & $0.517$ & $0.032$ & 5120 & $0.578$ & $0.035$ \\
 & All Active & 145 & $0.495$ & $0.023$ & 182 & $0.562$ & $0.025$ \\
 & Active-Not-Selective & 66 & $0.433$ & $0.028$ & 138 & $0.467$ & $0.031$ \\
 & Selective-Union & 140 & $0.530$ & $0.023$ & 142 & $0.579$ & $0.024$ \\
 & Selective & 79 & $\mathbf{0.538}$ & -- & 44 & $\mathbf{0.592}$ & -- \\
\bottomrule
\end{tabular}
\caption{Ordinal probe agreement with the median expert rating, for feature sets taken from layer 53, with each judge's decoded score as the \emph{Judge (Decoded)} baseline.
Every $\alpha$ is Krippendorff's $\alpha$ on the 320 test summaries.
Higher is better, and the best value per criterion per judge is bolded.
\emph{Selective-Union} pools the four criteria's selective sets, and its N falls well below their sum because many features are selective for several criteria at once.
Rows run from the judge's own output to the most selective feature set, so the \emph{Selective} reference row lies at the foot of each block.
For each comparison row, $\Delta = \alpha_{\mathrm{Selective}} - \alpha_{\mathrm{row}}$, so positive values favor \emph{Selective}.
sd$(\Delta)$ is the standard deviation of this paired difference over 10{,}000 bootstrap resamples of the test summaries. We score both predictors on the same resample with fitted models held fixed.
It describes uncertainty in the difference, not in either $\alpha$ alone.
We omit sd$(\Delta)$ for the \emph{Selective} reference row, whose difference from itself is zero.}
\label{tab:probing_results}
\end{table*}

\begin{figure}[t]
\centering
\includegraphics[width=\linewidth]{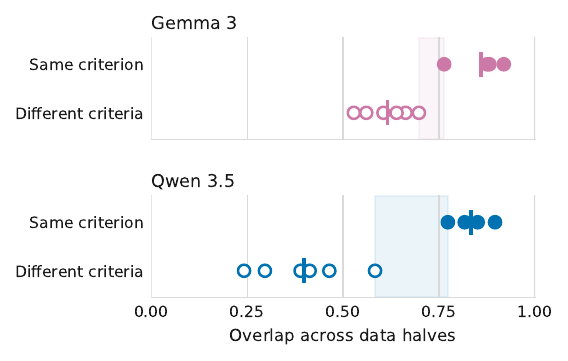}
\caption{Overlap between selective sets built on opposite halves of the training summaries.
Filled circles compare a criterion against itself, one per criterion, so they show how much the selection shifts with the sample alone.
They fall short of $1$ for that reason.
Open circles compare two different criteria, one per pair, and carry the same disagreement plus the criterion.
The vertical bar is each group's mean, and the shaded band is the gap between them.}
\label{fig:split_half}
\end{figure}

\subsection{Do Judges Distinguish Between the Four Criteria?}
\label{sec:exp2}

SummEval's two annotator pools differ on exactly this point.
The crowd rated the four criteria almost interchangeably while the experts separated them.
Asking which pattern a judge resembles is a way of asking how expert-like its internal representation is.

Each of the four criteria has its own selective set.
The question is whether these four feature sets are different.
If they are the same, that would imply that the models do not care about applying different rubrics or criteria and simply evaluate a summary based on some holistic quality.
But if they are different, that would imply that the models actually apply each criterion separately.
However, we cannot answer this question by simply comparing the feature sets.
This is because selection is based on a significance test (Section~\ref{sec:selectivity_method}), and a different sample of summaries would give us a somewhat different set of features that are labeled selective.
This means that two criteria will select different features even if the judge treats them identically.

We compute that difference by splitting the 1{,}280 training summaries into two halves of 640 and selecting features on each half separately.
A criterion's sets from the two halves already differ, since the halves are different samples.
We use the same split for all four criteria and compare coherence features from one half with consistency features from the other half. We measure this overlap using the overlap coefficient, which is the share of the smaller set that the two sets have in common, and average this value over 50 splits.

For two halves of the same criterion, the mean overlap is $0.86$ for Gemma~3 and $0.83$ for Qwen~3.5.
On the other hand, two halves from different criteria have a much lower overlap, $0.62$ and $0.40$, respectively.
In Figure~\ref{fig:split_half}, we see that for every pair of different criteria, the observed overlap is much lower than the overlap we observe when we compare a criterion with itself.
This means that it is the chosen criterion that separates the four sets, not the sample and not the number of features selected by each criterion (Appendix~\ref{app:permutation_null}).

Permuting the quality labels and re-selecting returns 2 to 5 features per criterion against the 42 to 87 we observe (Appendix~\ref{app:permutation_null}).
Therefore, the sets being compared are not merely a result of testing thousands of coordinates at once.
Similarly, the LLM judges' criterion-to-criterion score agreement is closer to that of the expert annotators than to that of the crowd-sourced annotators (Appendix~\ref{app:metric_agreement}).
However, neither judge is able to successfully reproduce the experts' grouping.

\subsection{Do Selective Feature Activations Beat the Judge's Decoded Score?}
\label{sec:probing_results}

The preceding results describe what is in a judge's activations.
But do these selective feature activations predict the expert label better than the judge's decoded score?
To answer this question, we fit the probes of Section~\ref{sec:probe_method} on five feature sets and compare them against the judge's own decoded score on the same 320 test summaries.

First we fix one LLM judge and one criterion $c$.
Let $A_c$ be the features active on the training summaries when the judge evaluates $c$ (Section~\ref{sec:background}), and let $S_c \subseteq A_c$ be those features we identified as selective for $c$ in Section~\ref{sec:selectivity_method}.
Now we can define five feature sets.
\emph{Selective} is $S_c$, the set of selective features for a particular criterion.
\emph{Selective-Union} is $\bigcup_{c'} S_{c'}$ over all four criteria, so it tests whether selecting per criterion has any advantage over computing selective features over all four criteria at once.
\emph{Active-Not-Selective} is $A_c \setminus S_c$, which is used to evaluate whether active features that were not labeled as selective have any predictive power when it comes to the expert judgment.
\emph{All Active} is $A_c$, the set of all active features, which is used to evaluate whether computing selective features even matters or just taking all active features would suffice.
\emph{Hidden State Activations} is the raw residual stream with no SAE at all, used as a baseline to measure whether we actually get anything out of using an SAE to extract features in the first place.

At the selected configurations, probes on selective features beat the judge's own decoded score in seven of the eight pairs (Table~\ref{tab:probing_results}), though both relevance margins fall inside two standard deviations.
The only exception is Gemma~3 on consistency, where the decoded score is higher by $0.082$.
One thing to note is that we do not claim that this result applies to other layers or widths.
In fact, for the earlier layers where there are too few selective features, it would be very difficult to train a meaningful probe.

For fluency, we observe the widest gap between the judge's decoded score and the predictions of the probes.
These scores disagree with the experts more than chance assignment would, at $-0.419$ and $-0.106$.
Probes fit on the activations that produced those very scores recover positive agreement, between $0.42$ and $0.60$.
Our probes are trained in a supervised manner on the expert ratings, whereas the decoded scores from the LLM judges are from a zero-shot setting.
We address that in Appendix~\ref{app:output_baselines}, where we fit a probe on the judge's decoded output.
That fitting improves the decoded score, but the probe on it still falls short of the probe on selective feature activations.
Surface statistics of the summary text explain none of this, reaching $0.138$ at best against the weakest probe's $0.42$ (Appendix~\ref{app:surface_baseline}).

Among the SAE-derived sets we see that the ordering \emph{Selective} $>$ \emph{Selective-Union} $>$ \emph{Active-Not-Selective} holds in all eight pairs.
We see that the margin over Selective-Union is small, but it never reverses.
Selective-Union strictly contains Selective. Despite that, it shows lower scores, implying that the added features from other criteria add more noise than useful information.
The margin over Active-Not-Selective is larger, so it is our selectivity metric rather than simply picking active features that identifies the informative features.
We see that with fewer than 90 selective features we are able to beat \emph{All Active} in all eight pairs.
Against the raw stream those selective sets also win in seven of eight, with small margins on both sides.
Therefore, a few dozen selective features predict expert ratings as well as the full residual stream, at under $2\%$ of its dimensionality.

\section{Discussion}
\label{sec:discussion}
An output-level meta-evaluation assesses a judge through its decoded score alone, and we go further by analyzing its internal activations.
Fluency shows the widest gap.
Both judges' decoded fluency scores fall below chance agreement with the experts, while probes on their activations agree positively.
This pattern gets repeated across other criteria like coherence and relevance, though the gap is not as stark.
Supervision alone cannot explain this pattern, since Appendix~\ref{app:output_baselines} gives the decoded score the same fitting and a gap remains.
Therefore, meta-evaluation that uses only the output understates what a judge represents.

SummEval has two annotator pools that differ on whether the four criteria are four things or one, and both judges fall closer to the experts than to the crowd~\citep{fabbri-etal-2021-summeval}.
Each criterion has its own selective set, and the split-half comparison of Section~\ref{sec:exp2} separates that result from sampling noise.
The decoded scores from both LLM judges point the same way (Appendix~\ref{app:metric_agreement}).
Neither judge groups the criteria as the experts do, so resembling the expert pattern is not the same as reproducing it.

Our feature selection also returns more flaw-identifying features than merit-identifying ones, in seven of the eight pairs.
One interpretation is that quality on these criteria is the absence of failure.
The asymmetry may be due to the SummEval benchmark instead, since three of its four criteria are written as instructions to deduct~\citep{fabbri-etal-2021-summeval}.
Or it may be due to our selectivity test.
Three of the four criteria have far more high-scoring summaries than low-scoring ones, so a feature that identifies flaws can reach a larger Cohen's $h$ than a feature that identifies merits in a summary (Appendix~\ref{app:selectivity_test}).
Coherence supports the third of these interpretations, since its groups are nearly balanced and the asymmetry weakens for Gemma~3 and reverses for Qwen~3.5.
On the other hand, relevance does not support this third interpretation, since both judges identify more flaws than its base rate allows.
A rubric written in positive terms would allow us to see if the judge still focuses on negative aspects of a summary when the instructions are not phrased as assigning penalties, and we leave that test to future work.

\section{Related Work}
\label{sec:related}

This work lies between two different fields.
One characterizes judges by what they output.
The other characterizes what models represent, but has not asked what an evaluator represents.

\textbf{LLM-as-a-judge evaluation} was established at scale by \citet{zheng2023judging}~\citep{gu2025surveyllmasajudge,li-etal-2025-generation}, though LLM ratings can successfully substitute for human ratings on some tasks and not others~\citep{chiang-lee-2023-large}.
G-Eval formalized single-output scoring~\citep{liu-etal-2023-g} and Prometheus~2 trains open evaluators against explicit rubrics~\citep{kim-etal-2024-prometheus}, with agreement varying sharply by task and criterion~\citep{bavaresco-etal-2025-llms}.
A large body of work then audits how judges fail.
Judges have been found to be sensitive to candidate order~\citep{wang-etal-2024-large-language-models-fair,shi-etal-2025-judging}.
They often reward style over substance~\citep{wu-aji-2025-style}, favor their own generations~\citep{panickssery2024llm}, try to identify which judgment a user will agree with rather than judging an item on its own merits~\citep{sharma2024sycophancy}, and disagree with their own earlier ratings~\citep{stureborg2024inconsistent,haldar-hockenmaier-2025-rating}.
Forcing one rating where several are defensible biases the protocol itself~\citep{validating_llm_judge}.
Every one of these diagnoses is made from the outputs, which is the constraint our results are about.

\textbf{Summarization metrics} moved from lexical overlap~\citep{lin-2004-rouge} to embedding comparison~\citep{zhao-etal-2019-moverscore,zhang2020bertscore} and then to metrics trained on human ratings~\citep{sellam-etal-2020-bleurt,rei-etal-2020-comet,yuan2021bartscore,zhong-etal-2022-towards}.
The annotations underneath proved unstable~\citep{kryscinski-etal-2019-neural,bhandari-etal-2020-evaluating}, and summarizers assert unsupported content~\citep{maynez-etal-2020-faithfulness}.
That prompted factual-consistency metrics built on weakly supervised classification~\citep{kryscinski-etal-2020-evaluating}, question generation and answering~\citep{wang-etal-2020-asking}, or sentence-level entailment~\citep{laban-etal-2022-summac}, which is how our consistency criterion is defined.
Recent studies use LLM judges to evaluate summarization directly~\citep{zhang-etal-2024-benchmarking,song-etal-2024-finesure}, some using them to evaluate several aspects at once~\citep{jain-etal-2023-multi,liu-etal-2024-x}.
\citet{wang-etal-2023-chatgpt} report per-criterion SummEval correlations for an LLM judge, the closest published comparison for ours.
We take the median of the expert ratings as the ground truth.
NLG evaluations are also reported too inconsistently to compare~\citep{howcroft-etal-2020-twenty}.
Chance-corrected agreement is contested for ordinal NLG ratings~\citep{amidei-etal-2019-agreement}, and resampling gives such comparisons their intervals~\citep{deutsch-etal-2021-statistical,deutsch-etal-2022-examining}.

\textbf{Sparse autoencoders} are described in Section~\ref{sec:background}, and two further points bear on our design.
Matryoshka training nests dictionaries so that low-index features should be coarser~\citep{bussmann2025learning,matryoshka_rl}.
The features these methods recover are usually matched to concepts with a concrete meaning~\citep{templeton2026}, through explanations from top-activating examples~\citep{paulo2024automatically}.
An evaluation criterion is not that kind of target, so our selection is statistical rather than semantic.
SAE features also do not reliably beat simpler baselines on probing, steering or concept detection~\citep{kantamneni2025sparse,wu2025axbench}, which is why we report raw residual-stream probes alongside them.

\textbf{Linear probes} are the standard tool for asking what a frozen representation encodes~\citep{alain2017understanding,belinkov-2022-probing}.
However, an expressive probe can learn the task rather than recover it~\citep{hewitt-liang-2019-designing,elazar-etal-2021-amnesic}.
Unsupervised methods recover latent knowledge~\citep{burns2023discovering}, and probes extract it while a model asserts a falsehood~\citep{azaria-mitchell-2023-internal} and expose its causal linear structure~\citep{marks2023geometry,li2023inference}.
The same logic extends to other high-level behaviors~\citep{zou2023representation}.
Probes also serve as evaluators themselves~\citep{ch-wang-etal-2024-androids}, and models hold usable internal estimates of their own correctness~\citep{kadavath2022language}.
A model's verbalized confidence also diverges from what its own probabilities support~\citep{lin2022teaching,tian-etal-2023-just}.
\citet{yuchi-etal-2026-llms} report the closest prior form of our gap.
Linear probes recover numeral magnitudes from hidden states and order two numerals at over $90\%$ accuracy, while the same models verbalize that comparison at $50$ to $70\%$.
In their experiments the probes had to recover a fixed value that a model was given, whereas in our experiment the criteria are more abstract, with no fixed value in the prompt.
None of this work asks whether a model acting as an evaluator represents the criterion it has been handed.

\textbf{Interpretability of evaluators} is a small but growing line of work.
\citet{wedgwood2026automated} also apply SAEs to judges, but they observed features that were related to \emph{pairwise} preference, while our target is absolute scoring on a given scale against a given rubric.
\citet{li2026rethinking} and \citet{sun2026borp} likewise probe evaluator activations, but to replace the judge with something cheaper.
Our aim is to measure what information in an LLM judge's hidden activations gets lost in its decoded score.
This decoded score acts as a baseline in our experiments.

\section{Conclusion}
\label{sec:conclusion}

We asked what an LLM judge represents when it applies an evaluation criterion, and whether it is able to distinguish the four criteria the way an expert human evaluator does.
Applying sparse autoencoders to two LLM judges rating summaries from SummEval, we find that quality-selective features are few in number, are found mostly in the later layers, are specific to the criterion under consideration, and identify flaws more often than merits.
Probes fit on a few dozen of these features can predict expert ratings better than the judges' own decoded scores in seven of eight cases.
Additionally, these probes match the performance of probes trained on the full hidden state despite using under $2\%$ of the features.
On fluency they show positive agreement, whereas the decoded scores show negative agreement.
Together, our results show that meta-evaluation experiments that only study a judge's final output will be incomplete.
We offer such probes as a way to diagnose a judge, not as a replacement for its score. This diagnosis needs only an open-weight judge, a released SAE, and expert ratings.

\section*{Limitations}

Our findings describe two open-weight judges of similar size scoring English news summaries on four published criteria, with activations recorded at one layer and the final prompt token.
By keeping the template fixed, we were able to compare the behavior of LLM judges on identical inputs. But this came at the cost of not being able to measure the impact of our phrasing in our prompts on a judge's application of an evaluation criterion.
Recording one position keeps score and activations in one run, at the cost of excluding features that fire while the judge processes the article.
Extending the sweep to every layer is a direction for future work, alongside extensions to other criteria and tasks~\citep{bavaresco-etal-2025-llms}.

Our evidence is correlational by design~\citep{gao-etal-2024-insights,orgad2025llms}.
It does not show that a judge actually consults these features when it generates a score~\citep{hewitt-liang-2019-designing,elazar-etal-2021-amnesic}.
Interventions such as steering, already demonstrated on a related evaluator~\citep{zhou2026verifysteer}, would help us answer that question.

SummEval's 16 systems predate current LLM summarizers, so the summaries carry more surface errors than contemporary output would.
Our split is stratified on the expert rating but not grouped by article or system (Section~\ref{sec:splits}), so a probe could recover either identity in place of quality.
In a one-way analysis of variance, article identity explains almost none of the expert rating variance and system identity at most $0.31$.
A probe recovering system identity would predict all four criteria about equally well, yet per-criterion selection wins in all eight pairs (Table~\ref{tab:probing_results}).
Surface statistics of the text, which carry system style, fail to predict the expert ratings (Appendix~\ref{app:surface_baseline}).

It is also an open question how faithfully SAE features describe the activations they decompose~\citep{sharkey2025open,karvonen2025saebench}.
Because of this, we report which features are selective and what they predict, rather than what they encode.

\section*{Ethical Considerations}

This work analyzes two existing open-weight models and one existing public dataset.
We do not train any new models and we do not collect any new human annotations, so our compute cost is limited to inference and probe fitting.

The probes are supervised, so they inherit whatever their labels encode.
SummEval's expert ratings cover English news summaries rated by a small pool of annotators.
A probe fit on those labels reflects that standard, and should be validated on the text it is applied to before being used as a general-purpose quality metric.

\section*{Acknowledgments}
This work used the Delta advanced computing and data resource. The National
Science Foundation (award OAC 2005572) and the State of Illinois support
Delta. Delta is a joint effort of the University of Illinois Urbana-Champaign
and its National Center for Supercomputing Applications.

\bibliography{custom,anthology-1,anthology-11,anthology-2}

\appendix

\input{appendix}

\end{document}

%% file: appendix.tex
\section{Appendix}
\label{sec:appendix}

\subsection{Artifacts and Licenses}
\label{app:licenses}

We use existing artifacts only and do not release any new data.
SummEval's annotations are distributed under the MIT License~\citep{fabbri-etal-2021-summeval}. The underlying CNN/DailyMail articles are distributed under Apache~2.0.
Gemma-3-27b-it is released under the Gemma Terms of Use~\citep{gemma3technicalreport}, a custom Google license that is not OSI-approved. Qwen-3.5-27B is released under Apache~2.0~\citep{qwen35technicalreport}.
Of the two SAE suites, Gemma Scope~2 is released under CC-BY-4.0~\citep{gemmascope2} and Qwen-Scope under Apache~2.0~\citep{deng2026qwenscopeturningsparsefeatures}.
Our analysis builds on SAELens (MIT)~\citep{saelens}, scikit-learn (BSD 3-Clause)~\citep{pedregosa2011scikit}, and mord (BSD 3-Clause)~\citep{pedregosa2017consistency}.

Every artifact is used for non-commercial academic research.
That is consistent with the intended use each license specifies, and with SummEval's release as a benchmark for studying summarization evaluation.
We do not add any annotations or redistribute any data.

\paragraph{Reproducibility.}
Our code is publicly available at \url{https://github.com/rajarshihaldar/probing-summary-judges}.
In Appendix~\ref{app:sae-setup} we share all the SAE release identifiers and sparsity settings, as well as the extraction procedure.
In Appendix~\ref{app:judgments} we share the full prompts used in our experiments.
In Section~\ref{sec:selectivity_method} we define our selectivity metric that we use to identify selective features.
The train-test split of Section~\ref{sec:splits} uses random seed 42.
Feature extraction and probing ran in separate environments.
Extraction used SAELens 6.44.0 with PyTorch 2.12.0 and Transformers 5.9.0.
Probing used mord 0.7 with scikit-learn 1.8.0.
For computing agreement with Krippendorff's alpha, we used krippendorff 0.8.2, and for performing Fisher's exact test in Section~\ref{sec:selectivity_method}, we used SciPy 1.17.1.
In Appendix~\ref{app:compute} we report the hardware used for the experiments and the computational cost of each stage.

\subsection{Computational Budget}
\label{app:compute}

All experiments ran on the NCSA Delta cluster with four NVIDIA A100 40GB GPUs per node.
Each job requested all four GPUs and 64\,GB of system memory.
Both judges have roughly 27B parameters.
We use the judges and both SAE suites as released, so our computational cost consists of inference and probe fitting.
Scoring the 1,600 summaries on four criteria with both judges, and extracting SAE activations across the four layers and dictionary widths, took 314.4 A100 GPU-hours over 18 jobs.
That figure counts GPU time allocated rather than used, and it includes exploratory and failed runs.
Fitting the ordinal probes on CPUs takes minutes, a negligible cost compared with feature extraction.

\subsection{AI Assistance}
\label{app:ai-assistance}

We used AI assistants to critique drafts of this paper and help proofread and revise the text.

\subsection{SummEval Dataset Details}
\label{app:summeval_details}

Here, we describe SummEval~\citep{fabbri-etal-2021-summeval}, the benchmark used in our experiments.

This benchmark contains summaries generated by 16 neural summarization systems on CNN/DailyMail and expert and crowd-sourced ratings on four different quality dimensions (coherence, consistency, fluency and relevance) for each of those summaries.
We share its license in Appendix~\ref{app:licenses}.

\paragraph{Data composition.}
The human-annotation set covers 100 articles from the CNN/DailyMail test split, each paired with summaries from all 16 systems, giving $100 \times 16 = 1{,}600$ summaries.
Every summary is independently scored by 5 crowd-sourced workers and 3 expert annotators on all four quality dimensions, with no missing ratings.

\paragraph{Annotation protocol.}
Crowd-sourced annotators were recruited through Amazon Mechanical Turk.
Because non-expert judgments were known to diverge from expert ones, the authors also recruited three annotators with a background in summarization research to rate the same summaries independently under the same instructions.
All annotators rated summaries on a 1-5 Likert scale (higher is better) along each criterion.
Each group included the dataset's reference summary, which provided a common point of comparison.

\paragraph{How the criteria are phrased.}
The full definitions of three criteria contain instructions for identifying errors that should lower the score~\citep{dang-2006-duc,fabbri-etal-2021-summeval}.
Consistency penalizes hallucinated facts, relevance penalizes redundant or excess content, and fluency is defined by the absence of formatting problems, capitalization errors and ungrammatical constructions.
Only coherence is stated positively: sentences should form a unified body of information rather than a collection of loosely related statements.
In Section~\ref{sec:discussion}, we discuss whether these definitions help explain why more selective features identify flaws.

\paragraph{Annotation quality and agreement.}
The SummEval authors report Krippendorff's $\alpha$ of 0.4920 for crowd-sourced annotations and 0.4132 for the first round of expert annotations, rising to 0.7127 after a second, corrective expert round.
These values are not directly comparable with ours.
SummEval used interval distances and compared individual annotations with one another.
We use ordinal distances and compare a scoring source with the expert median (Appendix~\ref{app:expert_ceiling}).
Crowd-sourced scores were found to be highly correlated across the four criteria, which suggests that crowd workers apply a single holistic notion of quality.
The expert ratings, on the other hand, were found to be able to distinguish the four criteria more clearly.

\subsection{Obtaining LLM Judgments From SummEval}
\label{app:judgments}

\paragraph{Record structure.}
We use the aligned, paired release of the SummEval annotations, which contains exactly 1,600 records, one per (article, system summary) pair.
Each record carries eight fields.
\texttt{text} is the source article and \texttt{decoded} the system summary.
These are the only two fields we feed to the judge.
\texttt{expert\_annotations} stores one score dictionary for each of three experts.
Each dictionary contains scores for coherence, consistency, fluency and relevance.
We use the median expert score for each criterion as the ground-truth label throughout.
\texttt{turker\_annotations} holds five crowdsourced ratings in the same format, which we use only for the distributional comparison in Appendix~\ref{app:judgment_statistics}.
\texttt{references} holds 11 reference summaries, \texttt{model\_id} identifies the summarization system, and \texttt{id} and \texttt{filepath} locate the source story.
Figure~\ref{fig:summeval_record} shows one record in full.

\paragraph{Prompt construction.}
Our prompts follow G-Eval~\citep{liu-etal-2023-g}, which evaluates SummEval along these same four dimensions.
Each prompt uses G-Eval's form-filling format: a task introduction, one criterion, evaluation steps, the text to evaluate, and a form requesting a score.
We take the criteria verbatim from SummEval's definitions of the four dimensions~\citep{fabbri-etal-2021-summeval}, so the judge is given the same wording the expert annotators were.
We add a rubric that defines all five rating levels.
We use a separate prompt for each criterion, so each summary receives four independent judgments.
In Figure~\ref{fig:prompt_coherence} we show the coherence prompt verbatim.
In Figure~\ref{fig:prompt_rubrics} we show the criteria, rubrics and steps for consistency and relevance.
In Figure~\ref{fig:prompt_fluency} we show these same components for fluency.

\paragraph{Obtaining a score from the judge.}
We compute the rating with G-Eval's probability-weighted procedure rather than sampling a response and extracting a number.
At the first position after the prompt we take the logits for the five tokens \texttt{"1"} through \texttt{"5"}.
We renormalize them with a softmax over those five tokens and compute the expectation $\sum_{s=1}^{5} s \cdot P(s)$, and then round it to the nearest integer.
We also record the hidden state activations of all the layers at the last input token position when getting the score.

\begin{figure*}[ht]
\centering
\lstcaptionbar{accRecord}{SummEval record}
\begin{lstlisting}[style=recordlisting]
(*@\lstrechead{id}@*)         : dm-test-8764fb95bfad8ee849274873a92fb8d6b400eee2
(*@\lstrechead{model\_id}@*)   : M11
(*@\lstrechead{filepath}@*)   : cnndm/dailymail/stories/8764fb95bfad8ee849274873a92fb8d6b400eee2.story

(*@\lstrechead{text (source article, truncated):}@*)
Paul Merson has restarted his row with Andros Townsend after the Tottenham midfielder was
brought on with only seven minutes remaining in his team's 0-0 draw with Burnley on Sunday.
'Just been watching the game, did you miss the coach? #RubberDub #7minutes,' Merson put on
Twitter. Merson initially angered Townsend for writing in his Sky Sports column that 'if
Andros Townsend can get in (the England team) then it opens it up to anybody.' Paul Merson had
another dig at Andros Townsend after his appearance for Tottenham against Burnley Townsend was
brought on in the 83rd minute for Tottenham as they drew 0-0 against [...]

(*@\lstrechead{decoded (system summary):}@*)
paul merson was brought on with only seven minutes remaining in his team 's 0-0 draw with
burnley . andros townsend scored the tottenham midfielder in the 89th minute . paul merson had
another dig at andros townsend after his appearance . the midfielder had been brought on to
the england squad last week . click here for all the latest arsenal news news .

(*@\lstrechead{expert\_annotations (3 annotators):}@*)
  {coherence: 2, consistency: 1, fluency: 4, relevance: 2}
  {coherence: 1, consistency: 1, fluency: 2, relevance: 1}
  {coherence: 1, consistency: 1, fluency: 3, relevance: 2}
(*@\lstrechead{median expert label used in this work:}@*)
  {coherence: 1, consistency: 1, fluency: 3, relevance: 2}

(*@\lstrechead{turker\_annotations}@*) : 5 further ratings, same four fields (not used here)
(*@\lstrechead{references}@*)         : 11 reference summaries (not used here)
\end{lstlisting}

\caption{One record from SummEval.
The summary attributes the substitution to Merson instead of Townsend, misdescribes the scoring event, and includes unrelated website text.
Experts assign median scores of 1 for consistency, 2 for relevance, and 3 for fluency.}
\label{fig:summeval_record}
\end{figure*}

\begin{figure*}[ht]
\centering
\lstcaptionbar{accPrompt}{Coherence prompt}
\begin{lstlisting}[style=promptlisting]
You will be given one summary written for a news article.
Your task is to rate the summary on one metric.
Please make sure you read and understand these instructions carefully.
Please keep this document open while reviewing, and refer to it as needed.

(*@\lsthead{Evaluation Criteria:}@*)
(*@\lsthead{Coherence (1-5)}@*) - the collective quality of all sentences. We align this
dimension with the DUC quality question of structure and coherence whereby
"the summary should be well-structured and well-organized. The summary
should not just be a heap of related information, but should build from
sentence to sentence to a coherent body of information about a topic."

(*@\lsthead{Rubric:}@*)
- 5: Excellently structured. Sentences flow naturally and build a unified,
     coherent body of information.
- 4: Mostly well-structured with only minor awkwardness that does not
     impede understanding.
- 3: Discernible structure but noticeable gaps in logical flow or abrupt
     transitions at one or two points.
- 2: Poorly organized. Multiple disconnected sentences or out-of-order
     ideas that make the passage hard to follow.
- 1: No coherent structure. Reads as a heap of unrelated fragments with
     no logical progression.

(*@\lsthead{Evaluation Steps:}@*)
1. Read the news article carefully and identify the main topic and key points.
2. Read the summary and compare it to the news article. Check if the summary
   covers the main topic and key points, and if it presents them in a clear
   and logical order.
3. Check whether individual sentences connect logically and build on each other.
4. Assign a score for coherence on a scale of 1 to 5 based on the Evaluation
   Criteria and Rubric above.

(*@\lsthead{Source Text:}@*)
(*@\lstslotname{\{\{Document\}\}}@*)

(*@\lsthead{Summary:}@*)
(*@\lstslotname{\{\{Summary\}\}}@*)

(*@\lsthead{Evaluation Form (scores ONLY):}@*)
- Coherence:
\end{lstlisting}

\caption{The coherence prompt, given verbatim.
\texttt{\{\{Document\}\}} and \texttt{\{\{Summary\}\}} are replaced with the \texttt{text} and \texttt{decoded} fields of the record.
The consistency and relevance prompts share this structure with their own criteria, rubrics and steps.
The fluency prompt omits the \texttt{Source Text} block entirely, since fluency is a property of the summary alone.}
\label{fig:prompt_coherence}
\end{figure*}

\begin{figure*}[ht]
\centering
\lstcaptionbar{accPrompt}{Criteria, rubrics and steps}
\begin{lstlisting}[style=promptlisting]
(*@\lsthead{Evaluation Criteria:}@*)
(*@\lsthead{Consistency (1-5)}@*) - the factual alignment between the summary and the
summarized source. A factually consistent summary contains only statements
that are entailed by the source document. Annotators were also asked to
penalize summaries that contained hallucinated facts.

(*@\lsthead{Rubric:}@*)
- 5: All claims fully supported by the article. No hallucinations or
     distortions of any kind.
- 4: Nearly all claims supported. At most one minor imprecision that does
     not constitute a factual error.
- 3: Most claims supported, but one clear factual distortion or unsupported
     claim is present.
- 2: Multiple unsupported or distorted claims present, undermining factual
     reliability.
- 1: Pervasive hallucinations or claims that directly contradict the article.

(*@\lsthead{Evaluation Steps:}@*)
1. Read the news article carefully and identify the main facts and details
   it presents.
2. Read the summary and compare it to the article. Check if the summary
   contains any factual errors or hallucinated facts not supported by the
   article.
3. Assign a score for consistency based on the Evaluation Criteria and
   Rubric above.

(*@\lsthead{Evaluation Criteria:}@*)
(*@\lsthead{Relevance (1-5)}@*) - selection of important content from the source. The
summary should include only important information from the source document.
Annotators were instructed to penalize summaries which contained redundancies
and excess information.

(*@\lsthead{Rubric:}@*)
- 5: Captures all the most important information. No redundant or peripheral
     details included.
- 4: Covers most key points but omits one minor important detail, or includes
     one slightly peripheral detail.
- 3: Covers some key points but misses one significant piece of important
     information, or includes notable filler content.
- 2: Misses multiple important points or is dominated by peripheral details.
- 1: Fails to capture the main point of the article, or is filled with
     irrelevant information.

(*@\lsthead{Evaluation Steps:}@*)
1. Read the summary and the source document carefully.
2. Compare the summary to the source document and identify the main points
   of the article.
3. Assess how well the summary covers the main points, and how much
   irrelevant or redundant information it contains.
4. Assign a relevance score from 1 to 5 based on the Evaluation Criteria
   and Rubric above.

\end{lstlisting}

\caption{Evaluation criteria, rubrics and evaluation steps for consistency and relevance, all given verbatim.
Each block replaces the corresponding coherence block in the template of Figure~\ref{fig:prompt_coherence}.
We quote the criterion definitions from SummEval~\citep{fabbri-etal-2021-summeval} and write the five-level rubrics ourselves.}
\label{fig:prompt_rubrics}
\end{figure*}

\begin{figure*}[ht]
\centering
\lstcaptionbar{accPrompt}{Fluency criterion, rubric and steps}
\begin{lstlisting}[style=promptlisting]
(*@\lsthead{Evaluation Criteria:}@*)
(*@\lsthead{Fluency (1-5)}@*) - the quality of individual sentences. Sentences in the summary
"should have no formatting problems, capitalization errors or obviously
ungrammatical sentences (e.g., fragments, missing components) that make the
text difficult to read."

(*@\lsthead{Rubric:}@*)
- 5: All sentences are well-formed and easy to read. No grammatical errors,
     formatting problems, or unnatural phrasing of any kind.
- 4: Nearly all sentences are fluent. At most one minor grammatical issue or
     slightly awkward phrasing that does not impede readability.
- 3: Some sentences are fluent but there are noticeable grammatical errors,
     fragments, or unnatural constructions in one or two sentences.
- 2: Multiple sentences contain clear grammatical errors or formatting problems
     that make parts of the summary difficult to read.
- 1: The majority of sentences are poorly formed, fragmented, or so
     ungrammatical that the summary is hard to understand.

(*@\lsthead{Evaluation Steps:}@*)
1. Read each sentence of the summary individually.
2. Check for grammatical errors, capitalization issues, formatting problems,
   fragments, or missing sentence components.
3. Check for unnatural phrasing that impedes readability.
4. Note how many sentences are affected and the severity of errors.
5. Based on your analysis and the rubric above, assign a score from 1 to 5.
\end{lstlisting}

\caption{The fluency criterion, rubric and evaluation steps, given verbatim.
This block replaces the coherence block in the template of Figure~\ref{fig:prompt_coherence}, and the fluency prompt drops the \texttt{Source Text} section, since fluency is a property of the summary alone.
We quote the criterion definition from SummEval~\citep{fabbri-etal-2021-summeval} and write the five-level rubric ourselves.}
\label{fig:prompt_fluency}
\end{figure*}

\clearpage
\subsection{Judgment Statistics on SummEval}
\label{app:judgment_statistics}

We compare the scores produced by four sources: the expert median, the crowd median, and the two LLM judges.

\begin{figure*}[ht]
\centering
\includegraphics[width=\linewidth]{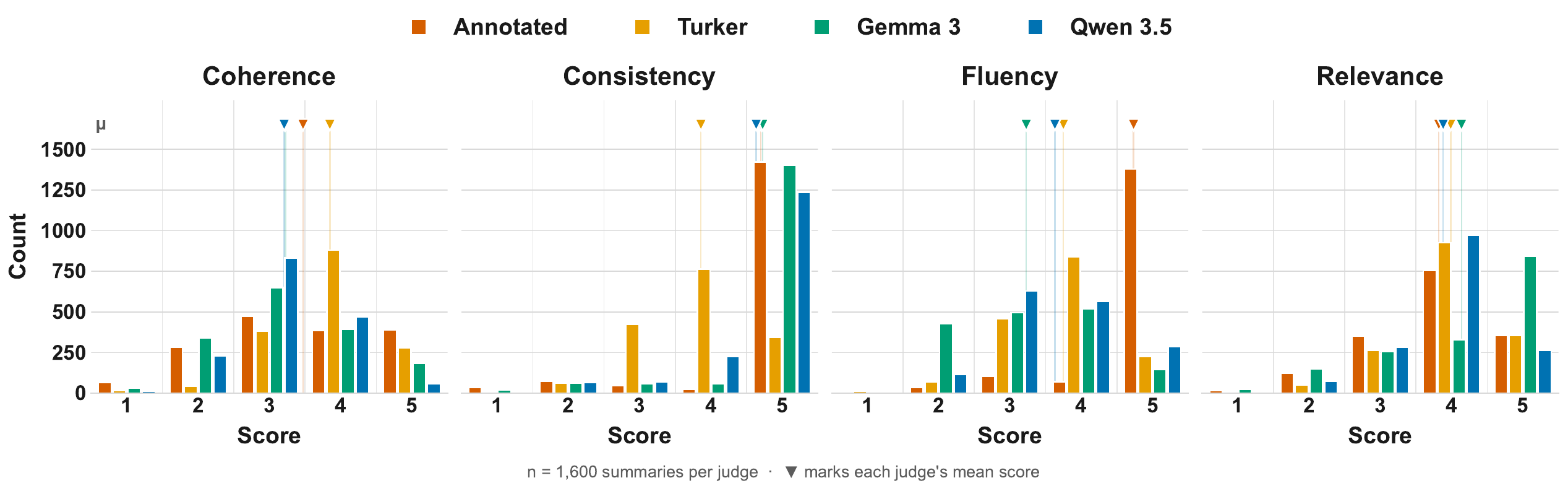}
\caption{Score distributions for the expert median (three annotations), crowd median (five annotations), Gemma~3, and Qwen~3.5.
Dashed lines indicate the mean score for each source.}
\label{fig:score_distributions}
\end{figure*}

\begin{figure*}[ht]
\centering
\begin{subfigure}{0.49\textwidth}
    \centering
    \includegraphics[width=\textwidth]{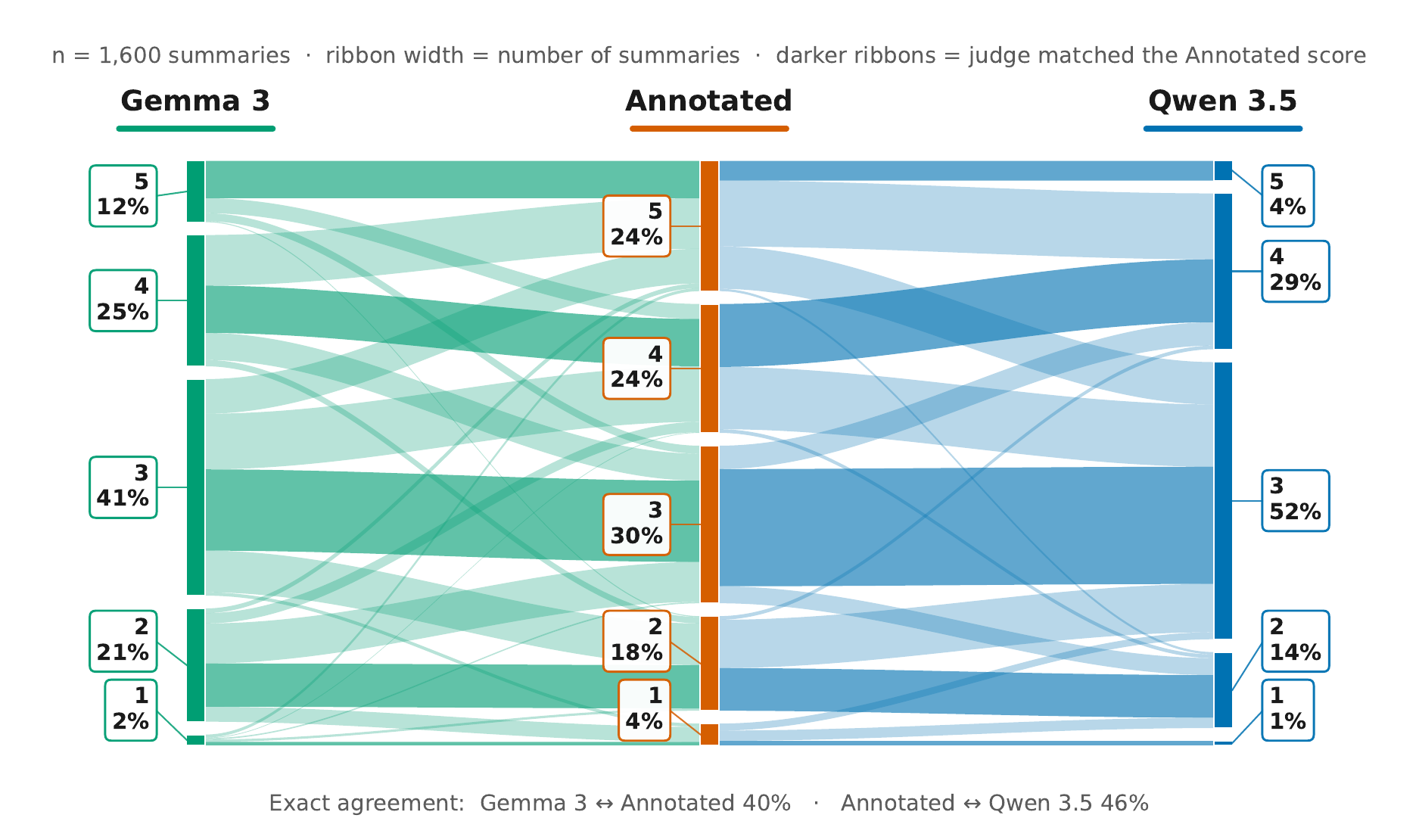}
    \caption{Coherence}
    \label{fig:sankey_coherence}
\end{subfigure}
\hfill
\begin{subfigure}{0.49\textwidth}
    \centering
    \includegraphics[width=\textwidth]{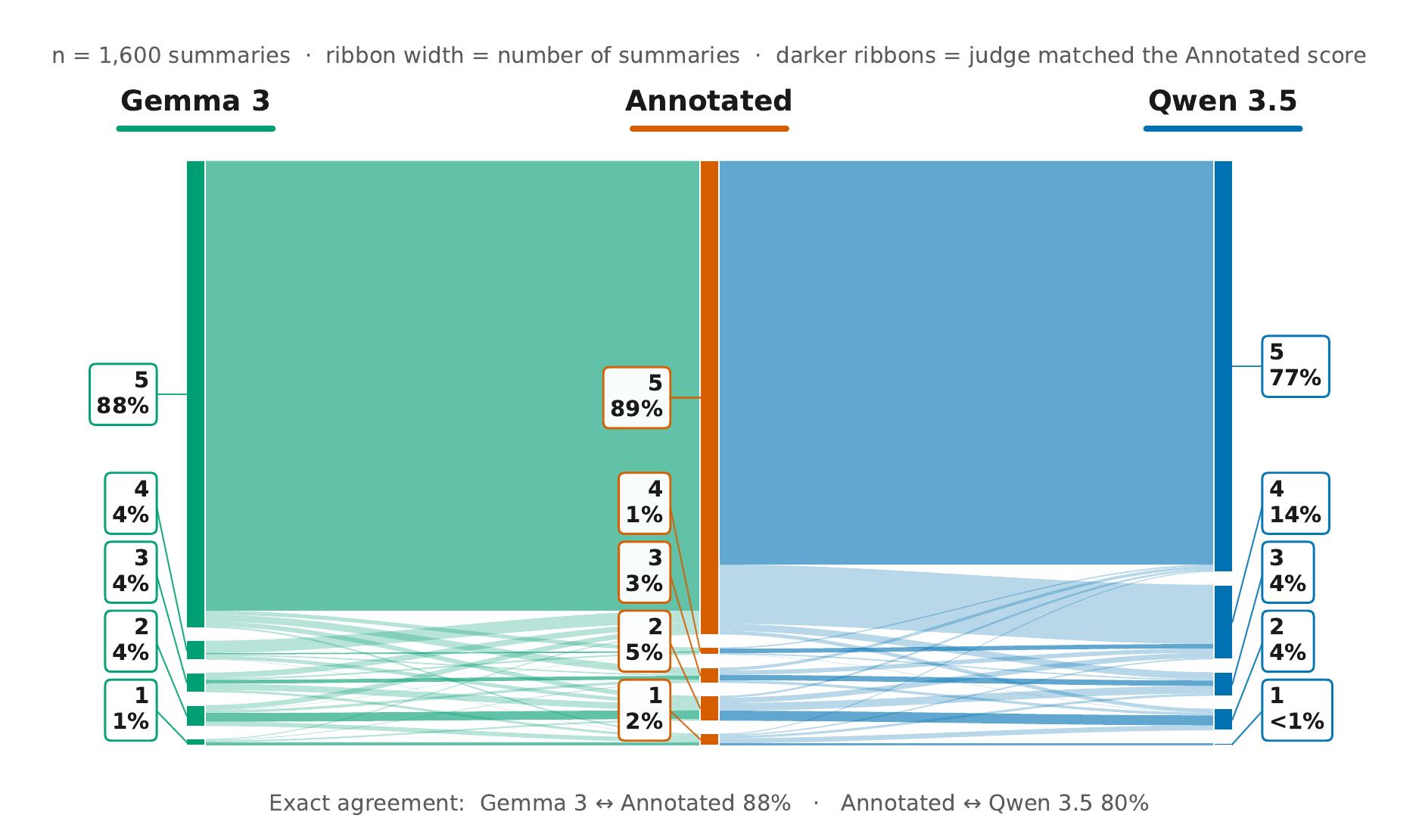}
    \caption{Consistency}
    \label{fig:sankey_consistency}
\end{subfigure}

\medskip
\begin{subfigure}{0.49\textwidth}
    \centering
    \includegraphics[width=\textwidth]{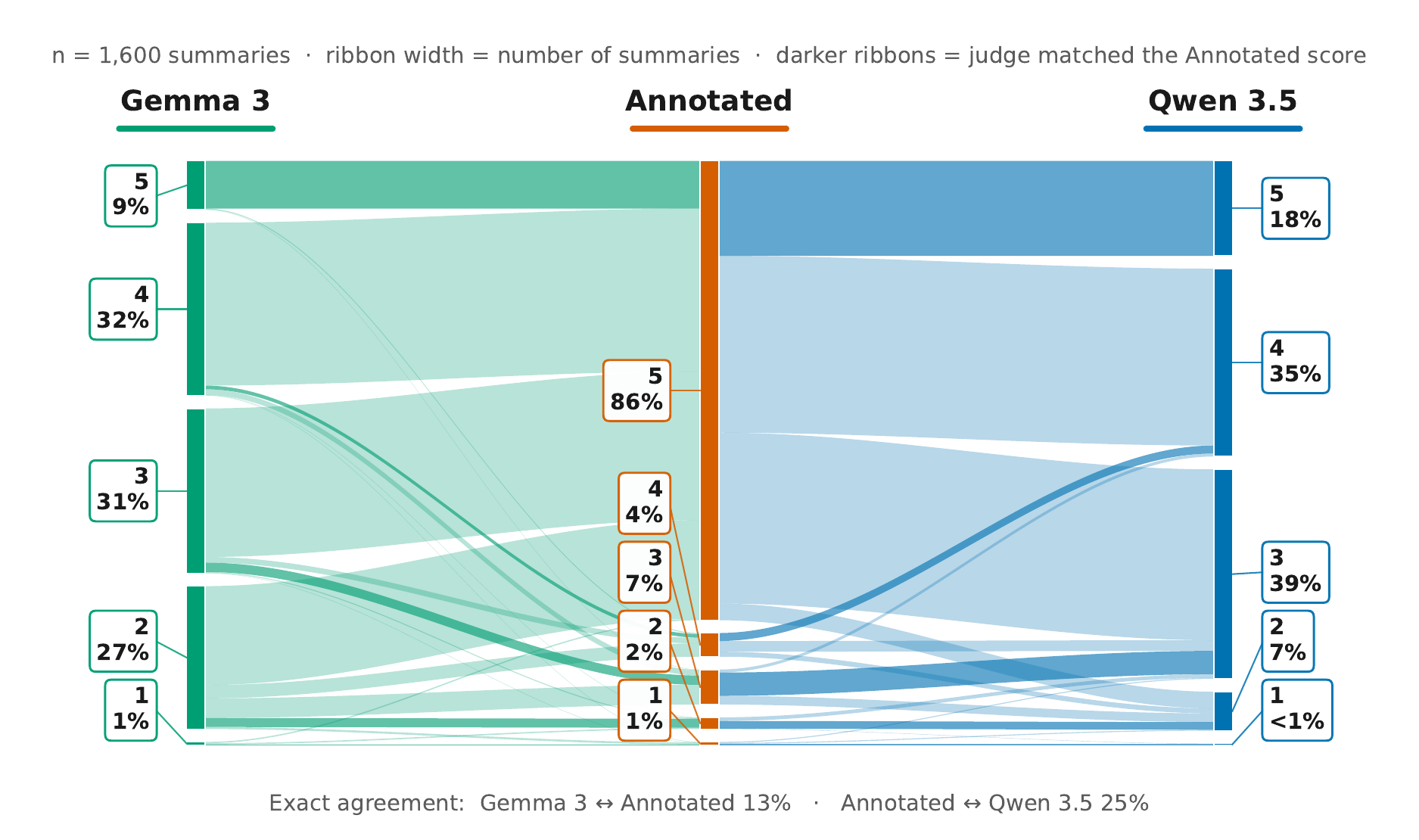}
    \caption{Fluency}
    \label{fig:sankey_fluency}
\end{subfigure}
\hfill
\begin{subfigure}{0.49\textwidth}
    \centering
    \includegraphics[width=\textwidth]{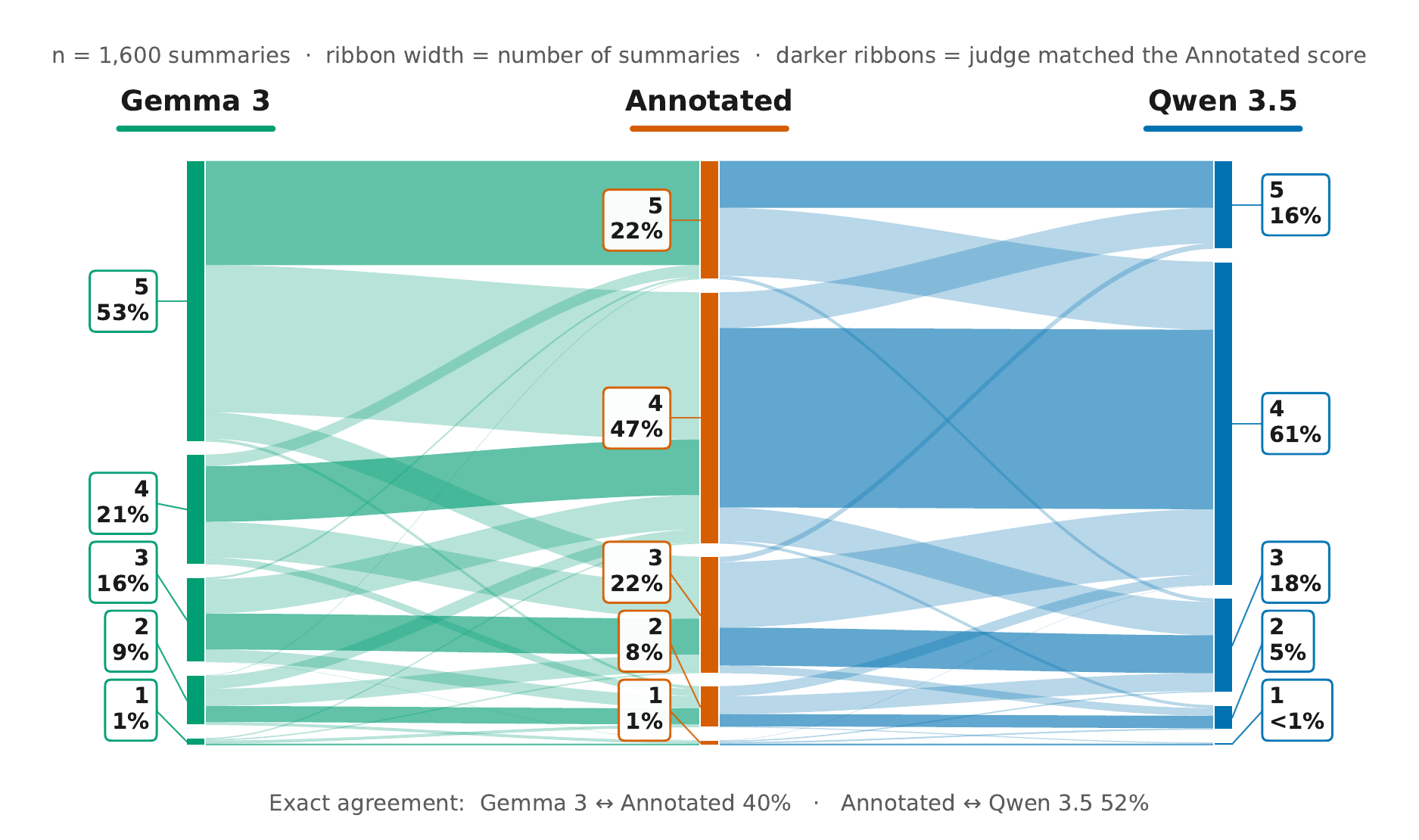}
    \caption{Relevance}
    \label{fig:sankey_relevance}
\end{subfigure}
\caption{Judge-to-expert score flow for each of the four SummEval criteria, over all 1,600 summaries.
In every panel the expert-annotated score is the center column, with Gemma-3's decoded score on the left and Qwen-3.5's on the right, so both judges are compared against a shared reference.
Each node shows the proportion of summaries assigned that score.
Ribbon width shows the number of summaries with each judge-expert score pair.
Saturated straight ribbons indicate identical judge and expert scores.
The footer of each panel gives the exact-agreement rates. Appendix~\ref{app:example_disagreement} compares the fluency rates with a modal-score baseline.}
\label{fig:sankey_all}
\end{figure*}

The analyses in this subsection use all 1,600 SummEval summaries.
These are exploratory checks that precede feature selection, so the train-test split of Section~\ref{sec:splits} plays no part in them.
The main-text Table~\ref{tab:judge_agreement} and decoded-score rows of Table~\ref{tab:probing_results} use the 320 test summaries so their baselines match the probe evaluation.
Appendix~\ref{app:full_agreement} gives their full-dataset counterpart.
The crowd annotations give a human but non-expert reference point throughout.

\subsubsection{Judge-Expert Agreement on the Full Dataset}
\label{app:full_agreement}

This table repeats the judge-expert agreement of Table~\ref{tab:judge_agreement} over all 1{,}600 summaries rather than the 320 test summaries.

\begin{table}[ht]
\centering
\begin{tabular}{lrr}
\toprule
Criterion & Gemma-3 & Qwen-3.5 \\
\midrule
Coherence   & 0.516  & 0.557  \\
Consistency & 0.648  & 0.546  \\
Fluency     & $-0.379$ & $-0.192$ \\
Relevance   & 0.472  & 0.509  \\
\bottomrule
\end{tabular}
\caption{Krippendorff's $\alpha$ (ordinal) between each judge's decoded score and the expert-annotated score over all 1,600 summaries.
In Table~\ref{tab:judge_agreement} we report the same agreement on the 320 held-out test summaries.}
\label{tab:judge_agreement_full}
\end{table}

Table~\ref{tab:judge_agreement_full} preserves the three main findings of the test-set comparison.
Each judge ranks the four criteria in the same order, fluency agreement is negative for both judges, and Qwen~3.5 leads Gemma~3 on every criterion except consistency.
The values differ by under $0.04$ $\alpha$ everywhere except Qwen~3.5 on fluency ($-0.106$ held-out against $-0.192$).

\clearpage
\subsubsection{Agreement Between Scoring Sources}
\label{app:source_agreement}

\begin{figure*}[ht]
\centering
\includegraphics[width=\textwidth]{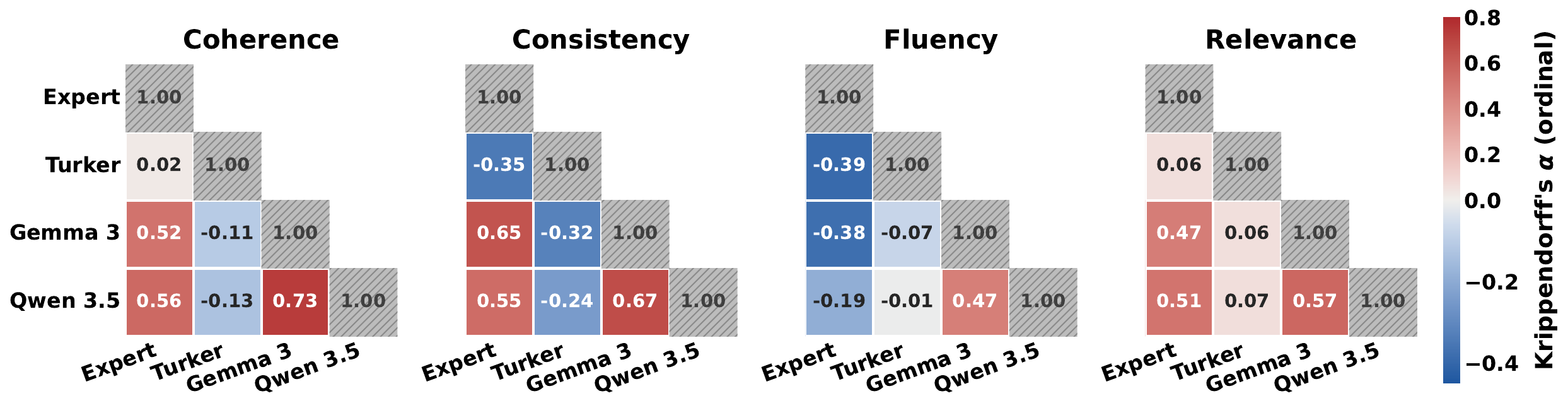}
\caption{Ordinal Krippendorff's $\alpha$ between every pair of scoring sources, one panel per criterion, over all 1,600 summaries.
The sources are the median of three expert annotations, the median of five crowd annotations, and each judge's decoded score.
The matrix is symmetric, so only the lower triangle is drawn.
The hatched diagonal is self-agreement.
The expert column reproduces the judge-expert agreement values in Table~\ref{tab:judge_agreement_full} and adds agreement between the crowd median and expert median.}
\label{fig:source_agreement}
\end{figure*}

We also compare the crowd median rating with expert ratings to examine whether low agreement is specific to LLM judges.
In Figure~\ref{fig:source_agreement} we report ordinal $\alpha$ between every pair of sources, with one panel per criterion.

The crowd median has much lower agreement with the experts than either judge on coherence, consistency, and relevance.
It reaches $\alpha = 0.02$ with the experts on coherence and $0.06$ on relevance, both close to zero, and is negative on consistency ($-0.35$) and fluency ($-0.39$).
Both judges beat it on three of the four criteria, by between $0.41$ and $0.99$ $\alpha$.
These values support the interpretation in Appendix~\ref{app:metric_agreement}.
The crowd does not separate the four criteria, and it does not track the experts on any of them either.

On all criteria except fluency, both judges agree more with the experts than with the crowd-sourced workers.
Subtracting each judge's agreement with the crowd from its agreement with the experts gives $+0.63$ and $+0.69$ on coherence, $+0.97$ and $+0.79$ on consistency, and $+0.41$ and $+0.44$ on relevance, for Gemma~3 and Qwen~3.5 respectively.

Fluency is the exception, where the LLM judges as well as the crowdsourced workers have negative agreement with the experts.
The crowd reaches $-0.39$, Gemma~3 $-0.38$, and Qwen~3.5 $-0.19$.
Probes fit on the judges' own activations recover $\alpha$ between $0.42$ and $0.60$ on that same criterion (Section~\ref{sec:probing_results}).

\subsubsection{Distribution of Scores}
\label{app:score_distributions}

Figure~\ref{fig:score_distributions} shows the distribution of scores on the 1-5 scale for each source and criterion.
The annotated scores are very high for both consistency and fluency, so the generated summaries are typically fluent and seldom contradict the source article.
Both judges assign high consistency scores, as the experts do, but assign much lower fluency scores.
Crowd ratings are more evenly distributed on both criteria.
A more even distribution does not imply better agreement with expert ratings.
On consistency the crowd sits at $\alpha = -0.35$ against the experts while Gemma~3 reaches $0.65$.

Two further patterns are visible in Figure~\ref{fig:score_distributions}.
The turkers place their modal score at 4 for every criterion, while the expert mode moves with the criterion.
Both judges' modal scores vary across criteria, as the experts' do.
Qwen~3.5's modal score matches the expert mode on three of the four criteria.
Even where their calibration is poor, the judges are more criterion-sensitive than the crowd workers.
Both LLM judges assign score 5 for coherence less often than the experts do.

Relevance shows most clearly why marginal distributions are not enough.
Mean relevance scores are similar across sources, but their distributions differ.
Gemma~3's modal score is 5, while Qwen~3.5 and the experts have mode 4.
Similar means do not show that the same summaries received the same scores.

\subsubsection{Example-Level Judge-Expert Disagreement}
\label{app:example_disagreement}

The median expert fluency rating was 5 for 86\% of the 1,600 summaries, as experts assigned perfect ratings to most summaries on fluency.
An LLM judge that always gave a 5 on fluency would therefore agree with the experts 86\% of the time.
Gemma~3 matched them on 13\% of summaries and Qwen~3.5 on 25\%.
In Figure~\ref{fig:sankey_fluency} we show that both judges usually scored summaries lower than the experts when they disagreed.
Even so, both judges' unrounded scores had a modest positive rank correlation with expert ratings (Kendall's $\tau_b$ of $0.251$ for Gemma~3 and $0.299$ for Qwen~3.5, Table~\ref{tab:kendall_tau}).

\begin{table}[ht]
\footnotesize
\centering
\begin{tabular}{lrrrr}
\toprule
& \multicolumn{2}{c}{Gemma-3} & \multicolumn{2}{c}{Qwen-3.5} \\
\cmidrule(lr){2-3} \cmidrule(lr){4-5}
Criterion & $\tau_b$ & $\alpha$ & $\tau_b$ & $\alpha$ \\
\midrule
Coherence   & $0.432$ & $0.516$   & $0.418$ & $0.557$ \\
Consistency & $0.393$ & $0.648$   & $0.375$ & $0.546$ \\
Fluency     & $0.251$ & $-0.379$  & $0.299$ & $-0.192$ \\
Relevance   & $0.438$ & $0.472$   & $0.408$ & $0.509$ \\
\bottomrule
\end{tabular}
\caption{Kendall's $\tau_b$ between each judge's decoded expectation and the median expert rating over all 1,600 summaries, beside the ordinal $\alpha$ of Table~\ref{tab:judge_agreement_full}.
$\tau_b$ uses the unrounded expectation, while $\alpha$ uses the rounded score. The statistics therefore use slightly different inputs.
$\tau_b$ varies little across criteria while $\alpha$ changes sign, which separates scale placement from ordering.}
\label{tab:kendall_tau}
\end{table}

\clearpage
\subsubsection{Inter-Criterion Agreement Within Each Scoring Source}
\label{app:metric_agreement}

Both judges separate the four criteria about as sharply as the experts do, and far more sharply than the crowd.
Their patterns of inter-criterion agreement differ from the experts' pattern.

\begin{figure*}[ht]
\centering
\includegraphics[width=\textwidth]{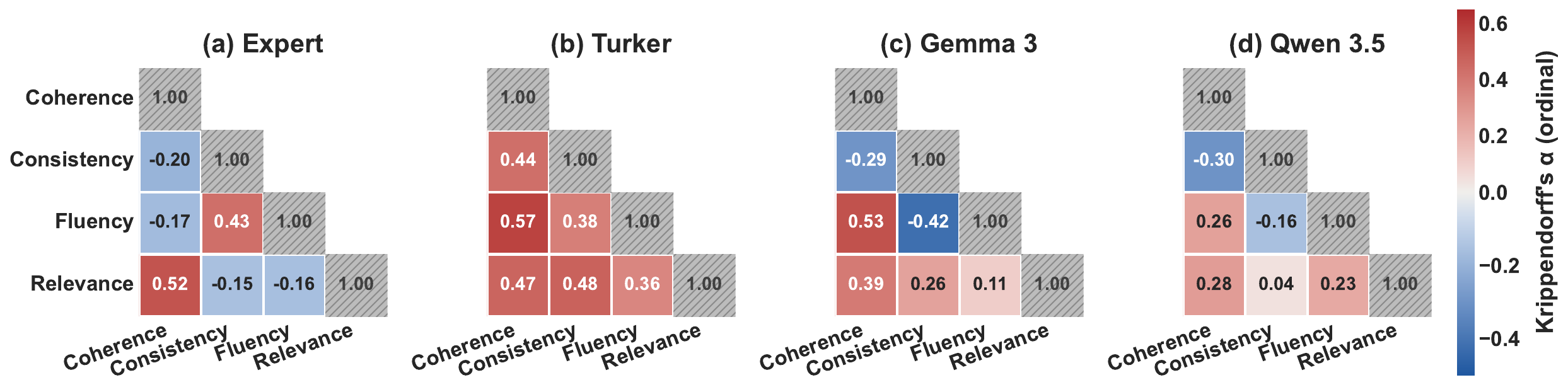}
\caption{Ordinal Krippendorff's $\alpha$ between every pair of SummEval criteria \emph{within} one scoring source: (a) the median of three expert annotations, (b) the median of five turker annotations, (c) Gemma~3, and (d) Qwen~3.5.
All four panels share one scale.
Each matrix is symmetric, so only the lower triangle is drawn.
The hatched diagonal is self-agreement.
Each cell treats the scores for two criteria as two raters of the same summaries, so a high value means that source did not distinguish those two criteria.}
\label{fig:metric_agreement_all}
\end{figure*}

Does a source treat the four criteria as four things or as one?
To answer it, we compute ordinal Krippendorff's $\alpha$ between every pair of criteria \emph{within} a single source, treating the two criterion score vectors as two raters of the same summaries.
A high value means that source's two scores rise and fall together, so it is not separating those criteria.
Figure~\ref{fig:metric_agreement_all} gives all four sources.

The experts divide the four criteria into two groups (Figure~\ref{fig:metric_agreement_all}a).
Coherence pairs with relevance ($\alpha = 0.52$) and consistency with fluency ($0.43$), while all four cross-group pairs are slightly negative, between $-0.20$ and $-0.15$.
Coherence and relevance concern what the summary selects and how it is organized.
Consistency and fluency concern whether it is faithful and well formed.
We do not observe any such pattern when it comes to turkers (Figure~\ref{fig:metric_agreement_all}b), with every pair between $0.36$ and $0.57$.
Averaged over the six pairs, off-diagonal $\alpha$ is $0.45$ for the turkers against $0.05$ for the experts.
The experts' mean is small because two large positive pairs cancel against four negative ones, and their mean $|\alpha|$ is $0.27$.
Averaging signed values partly cancels the positive and negative coefficients, so we report both the signed mean and mean absolute $\alpha$.
Therefore, the crowd annotations come close to scoring a single item four times.

Both judges' mean inter-criterion $\alpha$ values are close to the expert value.
Mean off-diagonal $\alpha$ is $0.10$ for Gemma~3 and $0.06$ for Qwen~3.5, so they distinguish the criteria about as sharply as the experts do, and more sharply than the crowd.
Mean $|\alpha|$ is $0.34$ for Gemma~3 and $0.21$ for Qwen~3.5, respectively above and below the experts' $0.27$.
The crowd reaches $0.45$ on both summaries, since none of its pairs is negative and nothing cancels.
Neither reproduces the experts' arrangement, though.
Gemma~3's strongest pair is coherence with fluency ($0.53$), which the experts place in different groups ($-0.17$), and its most negative is consistency with fluency ($-0.42$), which the experts pair ($0.43$).
Qwen~3.5 sets consistency apart from everything else ($-0.30$, $-0.16$, and $0.04$) and leaves the other three weakly related ($0.23$ to $0.28$).
Both judges distinguish the four criteria, but group them differently from the experts.

Inter-criterion score agreement and the feature-set overlap in Section~\ref{sec:exp2} show similar overall patterns but differ for individual criterion pairs.
The comparison is indicative rather than exact, since the $\alpha$ values here cover all 1,600 summaries while the selective sets come from the 1,280 training summaries.
Selective feature sets share little across criteria, and the decoded scores separate across criteria.
Qwen~3.5 has lower values on both measures, with selective Jaccard at most $0.28$ against Gemma~3's $0.56$, and mean off-diagonal $\alpha$ of $0.06$ against $0.10$.
We do not interpret that as a difference between the judges.
The dictionaries also differ in width and sparsity, which affect feature counts and prevent a direct comparison of the judges.
The criterion pairs with the most similar selective sets are not always those with the highest score agreement.
Gemma~3's least separated feature pair is coherence with relevance (Jaccard $0.56$), while its most agreeing score pair is coherence with fluency ($\alpha = 0.53$).
The two measures do agree on one pair.
Consistency with fluency gives Gemma~3's lowest selective overlap ($0.22$) and also its most negative score agreement ($-0.42$).
We take the mismatch as further evidence for the argument of Section~\ref{sec:discussion}.
A judge's decoded scores do not report its internal structure, so which criteria it separates internally cannot be inferred from which of its scores move together.

\subsection{A Matched Expert Ceiling}
\label{app:expert_ceiling}

Every $\alpha$ in this paper scores one source against the median of the three expert annotations.
\citet{fabbri-etal-2021-summeval} report $0.7127$ among their experts, but their method differs from ours.
They compare the three annotations to one another rather than to their median, and they use an interval distance metric while we use ordinal distance.
We therefore estimate an expert ceiling using our comparison method.
Each expert is scored against the median of the other two, using the same ordinal coefficient over the same 1{,}600 summaries.

\begin{table}[h]
\footnotesize
\centering
\begin{tabular}{lrrrr}
\toprule
Criterion & E1 & E2 & E3 & Mean \\
\midrule
Coherence & $0.542$ & $0.673$ & $0.587$ & $0.601$ \\
Consistency & $0.760$ & $0.760$ & $0.727$ & $0.749$ \\
Fluency & $0.589$ & $0.605$ & $0.519$ & $0.571$ \\
Relevance & $0.437$ & $0.377$ & $0.434$ & $0.416$ \\
\bottomrule
\end{tabular}
\caption{Leave-one-out expert ceiling.
Each column scores one expert against the median of the other two, using the ordinal Krippendorff's $\alpha$ of Section~\ref{sec:background}.}
\label{tab:expert_ceiling}
\end{table}

Across the three experts, mean agreement with the median rating of the other two experts was highest for consistency ($\alpha = 0.749$) and lowest for relevance ($\alpha = 0.416$).

There is one mismatch between these agreement values and the ones we report in Tables~\ref{tab:judge_agreement} and~\ref{tab:probing_results}.
An expert is scored against a median of two other raters, while every source we report is scored against a median of three.
A three-rater median has less rater noise and is therefore easier to predict, which makes this expert benchmark conservative.
A probe that reaches this value has not necessarily matched an individual expert's performance.

\subsection{Experimental Setup: SAE Feature Extraction}
\label{app:sae-setup}

We extracted sparse autoencoder (SAE) features from hidden-state activations of two instruction-tuned LLMs, Gemma-3-27B-it and Qwen-3.5-27B, using pretrained SAEs loaded via the \texttt{sae\_lens} library~\cite{saelens}.
Gemma-3-27B-it has 62 layers and a hidden dimension of 5376, and Qwen-3.5-27B has 64 layers and a hidden dimension of 5120.
For each model and example, we cached the hidden state at the last prompt token in each layer.
We encoded the selected states with the corresponding pretrained SAEs to obtain sparse feature representations.
We repeated this extraction for every layer and width listed in Section~\ref{sec:methodology}.

\paragraph{SAE configurations.}
For Gemma-3-27B-it, we used the Gemma Scope~2 residual-stream SAE (\texttt{gemma-scope-2-27b-it-res}), selecting individual SAEs by the identifier \texttt{layer\_\{N\}\_width\_\{W\}\_l0\_\{L\}} for $N \in \{16, 31, 40, 53\}$ and $W \in \{16\text{k}, 65\text{k}, 262\text{k}\}$, with target $L_0$ sparsity set to \emph{medium} (target $L_0$ in the $30$-$60$ range).
For Qwen-3.5-27B, we used the corresponding Qwen-Scope SAE (\texttt{qwen-scope-3.5-27b-w80k-l100}), selected by \texttt{layer\{N\}} over the same four layers, with dictionary width fixed at $80\text{k}$ and target $L_0$ sparsity of $100$.
The experiments of Section~\ref{sec:results} use layer 53 at width $16\text{k}$ for Gemma-3 and layer 53 at width $80\text{k}$ for Qwen-3.5.
Both SAEs were loaded via \texttt{SAE.from\_pretrained\_\allowbreak{}with\_cfg\_and\_sparsity}, which returns each SAE's trained weights, configuration and empirical sparsity statistics.

\paragraph{Release identifiers and resolved sparsity.}
The two releases resolve to the Hugging Face repositories \texttt{google/gemma-scope-2-27b-it} and \texttt{Qwen/SAE-Res-Qwen3.5-27B-W80K-L0\_100}.
We used revision \texttt{5c58dd4cddd5} for Gemma-3 extraction but did not record the Qwen revision.
The loader does not accept a revision argument, so a rerun uses the repository version available at that time.
The \emph{medium} setting specifies a sparsity range rather than a fixed value.
It resolves to a nominal $L_0$ of 53 at layer 16 and 60 at layers 31, 40 and 53, the same at all three widths.
Qwen-Scope is nominally $L_0 = 100$ at every layer.

\paragraph{Achieved sparsity on our prompts.}
Averaged over the 1{,}600 summaries at layer 53, Gemma~3 fires $43.2$ to $46.2$ features per prompt depending on the criterion, against that checkpoint's nominal $L_0$ of 60.
Qwen~3.5 fires exactly $100$ on every prompt, its target $L_0$, as its top-$k$ objective requires.

Those per-prompt figures should be compared against the union counts of Section~\ref{sec:feature_activation}.
The union across all 1{,}600 summaries contains 132 to 152 features for Gemma~3, roughly three times its per-prompt count, and 182 to 213 for Qwen~3.5, roughly twice its per-prompt count.
Different summaries therefore activate subsets of a relatively small feature pool.
The pool is not a fixed response to the prompt template, though.
Only 13 to 14 of Gemma~3's active features, between $9\%$ and $11\%$, fire on at least $99\%$ of summaries, and only 28 to 34 percent fire on even half of them.
A larger share of Qwen~3.5's active features fires on almost every summary, at $30\%$ to $38\%$.
This shows that most of each pool responds to the summary rather than to the rubric.
The counts of Section~\ref{sec:feature_activation} should still be interpreted as properties of this token position rather than of the judges in general.

\paragraph{Feature extraction procedure.}
Hidden-state activations have shape $(N, L+1, d)$, where $N$ is the number of examples, $L$ the number of layers, and $d$ the hidden dimension.
For each example, we selected the cached activation vector for the target layer, converted it to \texttt{float32}, and passed it through the SAE encoder.
We saved the resulting activations as PyTorch tensors indexed by model, criterion, layer, and dictionary width for downstream analysis.

\subsection{The Selectivity Significance Test}
\label{app:selectivity_test}

Section~\ref{sec:selectivity_method} requires Fisher's exact test to reject independence between a feature's firing and the quality label.

Write $N_{\text{high}}$ and $N_{\text{low}}$ for the numbers of high- and low-scoring training summaries, respectively, so that $N = N_{\text{high}} + N_{\text{low}} = 1{,}280$.
The stratified split of Section~\ref{sec:splits} fixes both in advance, which makes them design constants rather than estimates.
For a single feature, $f_{\text{high}}$ and $f_{\text{low}}$ count the summaries of each class it fires on, and $n = f_{\text{high}} + f_{\text{low}}$.

A binomial test on $p_{\text{high}}$ models those $n$ firings as independent draws from a pool that yields a high-scoring summary with probability $\pi_0$.
Two aspects of our design differ from that model.
Only $N_{\text{high}}$ high-scoring summaries exist, and a feature fires at most once on each, so the $n$ firings are a sample drawn \emph{without replacement} from the $N$ training summaries.
Conditioning on both margins gives the hypergeometric null:
\begin{equation}
\Pr(f_{\text{high}} = j) = \frac{\binom{N_{\text{high}}}{j}\binom{N_{\text{low}}}{n-j}}{\binom{N}{n}} .
\end{equation}
Its mean, $n\pi_0$, equals the binomial mean, but its variance includes the finite population correction $(N-n)/(N-1)$.
We compute the two-sided $p$-value using the minimum-likelihood rule: we sum $\Pr(f_{\text{high}} = j)$ over outcomes no more probable than the observed count, as \texttt{scipy.stats.fisher\_exact} does.

The choice only changes which features are selected when activation counts are small.
The correction factor is close to $1$ whenever $n \ll N$, so an asymptotic binomial test and the exact test agree except at small $n$.
When they disagree, the binomial test is the liberal one.
Take a feature that fires on five training summaries, all of them high-scoring, at $\pi_0 = 0.5$.
It gives $h = 1.57$ and is significant under the asymptotic binomial test, yet its exact two-sided $p$-value is $0.0625$.
The exact test excludes this feature despite its large effect size.

\paragraph{Alternatives compared.}
We compared five rules for selecting features. Three measured lift by comparing the share of a feature's firings on high-scoring summaries, $p_{\text{high}}$, with the base rate $\pi_0$. They used an absolute difference, a signed difference, or a ratio. The fourth rule used Cohen's $h$, and the fifth used a Bayesian method with its own significance criterion.

We evaluated each of the first four rules twice, once with a Wilson score interval and once with Fisher's exact test.
Cohen's $h$ with Fisher's test separated high- from low-scoring summaries best, so Section~\ref{sec:selectivity_method} adopts that pairing.

\paragraph{Class imbalance and attainable effect sizes.}
The high-scoring and low-scoring groups differ in size, and the imbalance varies by criterion.
On the 1{,}280 training summaries the expert median gives 622 high-scoring summaries against 658 low-scoring ones on coherence, 1{,}155 against 125 on consistency, 1{,}160 against 120 on fluency, and 889 against 391 on relevance.
The base rate $\pi_0$ runs from $0.486$ on coherence, the one criterion where low-scoring summaries are the majority, to $0.906$ on fluency.
The stratified split of Section~\ref{sec:splits} holds each of these within $0.0003$ of its value over all 1{,}600 summaries.

This variation affects the attainable range of Cohen's $h$ and the interpretation of Figure~\ref{fig:high_low}.
Cohen's $h$ is symmetric in the arcsine metric, but the range attainable at a given $\pi_0$ is not.
At coherence's $\pi_0 = 0.486$, a merit-identifying feature can reach $h = 1.60$ and a flaw-identifying feature $-1.54$, giving merit-identifying features a slightly larger attainable effect-size magnitude.
At fluency's $\pi_0 = 0.906$, the same bounds are $0.62$ and $-2.52$, so the attainable magnitude is four times larger for flaw-identifying features.
Fisher's power is likewise asymmetric when one margin is small.

We see that the excess of flaw-identifying features is smallest on coherence for Gemma~3, at $1.18$ per merit-identifying feature, and it reverses for Qwen~3.5, at $0.58$.

The other three criteria have imbalanced quality groups.
For consistency and fluency the observed ratios reach $3.14$ and $2.82$ for Gemma~3 and $1.59$ and $2.00$ for Qwen~3.5, all well below the roughly fourfold asymmetry the base rate permits.
Relevance reaches $2.59$ and $2.67$ against an allowance of $1.68$, so it is the one criterion where both judges identify flaws more often than the base rate alone would produce.
For Qwen~3.5 the ordering runs against the artifact account outright.
Its weakest asymmetry falls on consistency, the second most skewed criterion, and its strongest on relevance, the least skewed of the three.
Ranking the four criteria by $\pi_0$ against their flaw-to-merit ratios gives a Spearman $\rho$ of $0.80$ for Gemma~3 and $0.40$ for Qwen~3.5, on four points.

\paragraph{Multiple comparisons.}
We use Fisher's exact test to check whether a feature fires more often on high-scoring or low-scoring summaries.
We apply the 5\% significance threshold separately to each feature, without adjusting for the number of features tested.
The number of tests that can return a selection is much smaller than the size of the dictionary.
A feature that never fires gives a degenerate table and a $p$ of 1, so it cannot be selected at any threshold.
This means that the effective pool is the active set, 132 to 152 features per criterion for Gemma~3 and 182 to 213 for Qwen~3.5.
An uncorrected $\alpha = 0.05$ over a pool that size admits roughly 7 to 11 features under the null, against the 42-87 and 44-75 of Table~\ref{tab:selective_counts}.
Shuffling the quality labels and selecting again returns 2 to 5 features per criterion, so the measured rate sits below that bound (Appendix~\ref{app:permutation_null}).
The effect-size threshold rather than the test is still what bounds the selected sets.

\clearpage
\subsection{Three Checks on the Selective Sets}
\label{app:permutation_null}

\paragraph{Do the expert ratings predict which features pass the selectivity test?}
We keep each summary's feature activations fixed and shuffle the expert ratings across summaries.
We move all four ratings together to preserve the correlations between criteria.
For Gemma~3, the original ratings yield 42 to 87 selective features per criterion.
Across 199 shuffles, the average falls to 2 to 5.
Qwen~3.5 shows a similar drop (Table~\ref{tab:null_labels}).

\begin{table*}[t]
\footnotesize
\centering
\begin{tabular}{lrrrr}
\toprule
& \multicolumn{2}{c}{Gemma~3} & \multicolumn{2}{c}{Qwen~3.5} \\
\cmidrule(lr){2-3} \cmidrule(lr){4-5}
Criterion & Observed & By chance & Observed & By chance \\
\midrule
Coherence & $74$ & $1.9$ & $57$ & $3.1$ \\
Consistency & $87$ & $4.6$ & $75$ & $4.5$ \\
Fluency & $42$ & $4.4$ & $60$ & $5.7$ \\
Relevance & $79$ & $2.9$ & $44$ & $3.2$ \\
\midrule
Pooled & $140$ & $13.3$ & $142$ & $16.0$ \\
\bottomrule
\end{tabular}
\caption{Selective-feature counts under expert ratings and shuffled ratings.
The \emph{by chance} columns average 199 shuffles.
\emph{Pooled} counts the distinct features selected for any of the four criteria.}
\label{tab:null_labels}
\end{table*}

\begin{figure*}[ht]
\centering
\includegraphics[width=0.80\textwidth]{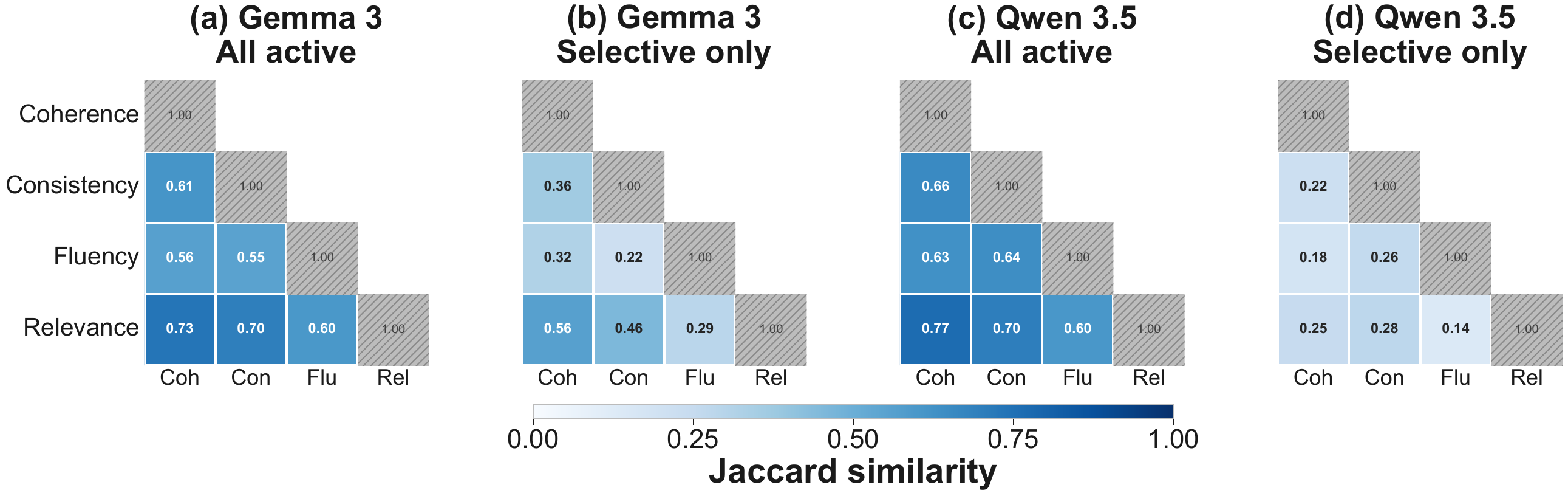}
\caption{Jaccard similarity across criterion pairs for all active features (a, c) versus \emph{selective} features only (b, d), for Gemma~3 (a, b) and Qwen~3.5 (c, d).
The matrix is symmetric, so only its lower triangle is drawn, and the hatched diagonal is self-overlap.
High values indicate that the criteria share many feature-set members.
Table~\ref{tab:null_sizes} compares the selective panels with random sets of the same sizes.}
\label{fig:jaccard}
\end{figure*}

\begin{figure}[h]
\centering
\includegraphics[width=\linewidth]{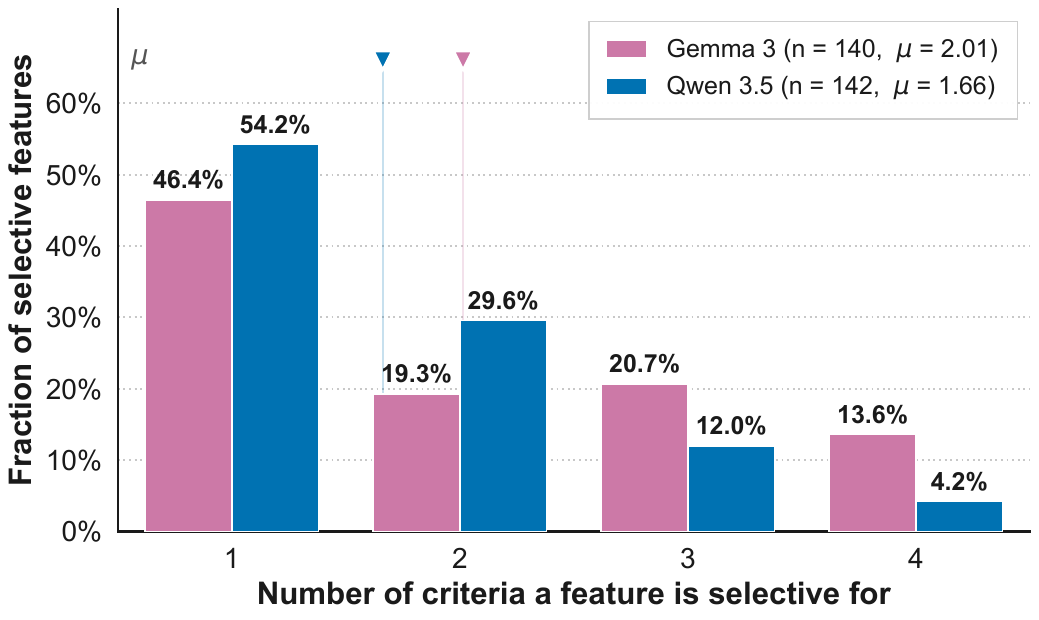}
\caption{Percentage of selective features that are selective for 1-4 criteria.
Single-criterion features are the largest share for both judges.
Triangles mark the mean number of criteria per selective feature for each judge, also given in the legend.
Table~\ref{tab:null_sizes} compares these means with the lower means from random sets of the same sizes.}
\label{fig:breadth_expert}
\end{figure}

\paragraph{Do different criteria share more features than expected by chance?}
For each criterion, we draw as many features from its active set as its selective set contains.
We repeat this 2{,}000 times and compare the Jaccard overlap of the random sets with the observed overlap (Table~\ref{tab:null_sizes} and Figure~\ref{fig:jaccard}).
In nine of the twelve comparisons across judges and criterion pairs, the observed overlap exceeds the $97.5$th percentile of the random draws.
Figure~\ref{fig:breadth_expert} shows how many criteria share each selective feature.

\begin{table*}[t]
\footnotesize
\centering
\begin{tabular}{lrrrr}
\toprule
& \multicolumn{2}{c}{Gemma~3} & \multicolumn{2}{c}{Qwen~3.5} \\
\cmidrule(lr){2-3} \cmidrule(lr){4-5}
Pair & Observed & By chance & Observed & By chance \\
\midrule
Coh, Con & $0.36$ & $0.26$ & $0.22$ & $0.15$ \\
Coh, Flu & $0.32$ & $0.16$ & $0.18$ & $0.13$ \\
Coh, Rel & $0.56$ & $0.29$ & $0.25$ & $0.13$ \\
Con, Flu & $0.22$ & $0.17$ & $0.26$ & $0.14$ \\
Con, Rel & $0.46$ & $0.30$ & $0.28$ & $0.13$ \\
Flu, Rel & $0.29$ & $0.18$ & $0.14$ & $0.11$ \\
\midrule
Criteria per feature & $2.01$ & $1.72$ & $1.66$ & $1.41$ \\
Pooled features & $140$ & $164$ & $142$ & $167$ \\
\bottomrule
\end{tabular}
\caption{Jaccard overlap between selective sets, compared with random subsets of the same sizes.
The \emph{by chance} columns average 2{,}000 draws, each taking as many features from a criterion's active set as that criterion actually selected.
Criteria are abbreviated to their first three letters.}
\label{tab:null_sizes}
\end{table*}

\paragraph{Does the criterion still matter when some features are shared?}
We split the 1{,}280 training summaries into two halves and select features independently on each half.
We measure what fraction of the smaller set also appears in the other set.
Across 50 splits, this overlap averages $86\%$ for Gemma~3 when both halves use the same criterion and $62\%$ when they use different criteria.
For Qwen~3.5, the corresponding values are $83\%$ and $40\%$.
For both judges, the within-criterion overlap remains higher after we subtract the overlap expected for random sets of the same sizes (Table~\ref{tab:split_half}).

\begin{table*}[t]
\footnotesize
\centering
\begin{tabular}{lrrrr}
\toprule
& \multicolumn{2}{c}{Gemma~3} & \multicolumn{2}{c}{Qwen~3.5} \\
\cmidrule(lr){2-3} \cmidrule(lr){4-5}
& Overlap & Above chance & Overlap & Above chance \\
\midrule
Same criterion & $0.86$ & $0.43$ & $0.83$ & $0.58$ \\
Different criteria & $0.62$ & $0.21$ & $0.40$ & $0.16$ \\
\midrule
Ratio & $0.72$ & $0.49$ & $0.48$ & $0.28$ \\
\bottomrule
\end{tabular}
\caption{Overlap between selective sets built on opposite halves of the training summaries, averaged over 50 splits.
\emph{Above chance} is that overlap minus what random sets of the same sizes reach when drawn from the same half's active features.
The ratio divides the second row by the first, so a value below $1$ means the criterion separates the sets by more than a change of sample does.}
\label{tab:split_half}
\end{table*}

\subsection{Baselines from the Judge's Output}
\label{app:output_baselines}

The activation probes in Section~\ref{sec:probing_results} learn from expert ratings.
The judge's decoded score comes directly from its output probabilities.
A probe could therefore improve on the decoded score simply by learning the experts' rating scale.
To test whether the activations offer any further advantage, we train two baselines on the same expert ratings using only the judge's output probabilities as inputs.

Two baselines do not require any training.
\emph{Rounded} gives the published score by rounding the expected rating to the nearest level.
\emph{Argmax} takes the most probable rating.
We fit the other two baselines on the same 1{,}280 expert-rated summaries as the probes.
\emph{Isotonic} learns a monotone mapping from the unrounded expected rating to the expert scale.
\emph{Probs} fits \texttt{LogisticAT} to all five rating probabilities.
For each baseline, we measure agreement with the median expert rating on the same 320 test summaries using ordinal Krippendorff's $\alpha$.

\begin{table}[h]
\footnotesize
\centering
\setlength{\tabcolsep}{3pt}
\begin{tabular}{llrr rr}
\toprule
& & \multicolumn{2}{c}{Gemma-3} & \multicolumn{2}{c}{Qwen-3.5} \\
\cmidrule(lr){3-4} \cmidrule(lr){5-6}
Criterion & Readout & $\alpha$ & sd$(\Delta)$ & $\alpha$ & sd$(\Delta)$ \\
\midrule
\multirow{5}{*}{Coherence}
 & Selective & $0.681$ & -- & $0.683$ & -- \\
 & Rounded & $0.500$ & $0.034$ & $0.590$ & $0.034$ \\
 & Argmax & $0.501$ & $0.035$ & $0.590$ & $0.036$ \\
 & Isotonic & $0.375$ & $0.032$ & $0.454$ & $0.033$ \\
 & Probs & $0.511$ & $0.030$ & $0.521$ & $0.030$ \\
\midrule
\multirow{5}{*}{Consistency}
 & Selective & $0.557$ & -- & $0.694$ & -- \\
 & Rounded & $0.639$ & $0.032$ & $0.532$ & $0.032$ \\
 & Argmax & $0.639$ & $0.032$ & $0.537$ & $0.033$ \\
 & Isotonic & $0.534$ & $0.030$ & $0.533$ & $0.030$ \\
 & Probs & $0.537$ & $0.030$ & $0.565$ & $0.028$ \\
\midrule
\multirow{5}{*}{Fluency}
 & Selective & $0.558$ & -- & $0.600$ & -- \\
 & Rounded & $-0.419$ & $0.040$ & $-0.106$ & $0.036$ \\
 & Argmax & $-0.483$ & $0.040$ & $-0.202$ & $0.038$ \\
 & Isotonic & $0.286$ & $0.037$ & $0.293$ & $0.035$ \\
 & Probs & $0.380$ & $0.036$ & $0.392$ & $0.032$ \\
\midrule
\multirow{5}{*}{Relevance}
 & Selective & $0.538$ & -- & $0.592$ & -- \\
 & Rounded & $0.480$ & $0.034$ & $0.521$ & $0.038$ \\
 & Argmax & $0.482$ & $0.033$ & $0.518$ & $0.038$ \\
 & Isotonic & $0.372$ & $0.032$ & $0.301$ & $0.034$ \\
 & Probs & $0.411$ & $0.032$ & $0.355$ & $0.032$ \\
\bottomrule
\end{tabular}
\caption{Ordinal agreement with the median expert rating for four baselines built from the judge's own output, on the 320 test summaries of Section~\ref{sec:splits}.
\emph{Rounded} and \emph{Argmax} require no fitting to expert labels.
\emph{Isotonic} and \emph{Probs} are fitted on the same 1{,}280 training summaries as the probes.
\emph{Selective} repeats the selective-feature probe of Table~\ref{tab:probing_results} as the reference row.
For each comparison row, $\Delta = \alpha_{\mathrm{Selective}} - \alpha_{\mathrm{row}}$, so positive values favor \emph{Selective}.
sd$(\Delta)$ is the standard deviation of this paired difference over 10{,}000 bootstrap resamples of the test summaries. We score both predictors on the same resample with fitted models held fixed.
It describes uncertainty in the comparison with \emph{Selective}, not between two output baselines.
We omit sd$(\Delta)$ for the \emph{Selective} reference row, whose difference from itself is zero.}
\label{tab:output_baselines}
\end{table}

Replacing the rounded expectation with \emph{Argmax} does not recover fluency agreement.
\emph{Argmax} tracks \emph{Rounded} within $0.005$ on every criterion except fluency, where it is worse on both judges.

For fluency, \emph{Isotonic} raises agreement with the experts from $-0.419$ to $0.286$ for Gemma~3 and from $-0.106$ to $0.293$ for Qwen~3.5.
\emph{Probs}, which uses all five score probabilities, reaches $0.380$ for Gemma~3 and $0.392$ for Qwen~3.5.
Using the judge's output alone closes most of the fluency gap.
The selective probe still exceeds \emph{Probs} by $0.178$ for Gemma~3 and $0.208$ for Qwen~3.5.

On coherence, consistency, and relevance, \emph{Isotonic} agrees less closely with the experts than \emph{Rounded} in five of the six comparisons.
The largest drop is on relevance for Qwen~3.5, from $0.521$ to $0.301$.
On consistency for the same judge, the two baselines give nearly identical results.
\emph{Probs} achieves higher agreement than \emph{Isotonic} on all four criteria for both judges.

For each judge-criterion pair, we compare the selective probe with the output baseline having the highest $\alpha$. The selective probe has higher agreement in seven of eight pairs.
The exception is Gemma~3 on consistency, the same pair Section~\ref{sec:probing_results} names.
The relevance differences are the smallest, at $0.056$ and $0.071$.
Both are smaller than twice their paired bootstrap standard deviations, $0.033$ and $0.038$.

\subsection{A Surface-Feature Baseline}
\label{app:surface_baseline}

We test whether a probe that uses only surface features of the summary can agree with the experts as closely as the activation probes do.

We use 13 surface features.
They are token count, sentence count, mean sentence length, type-token ratio, the fraction of uppercase alphabetic characters, the fraction of sentence-initial tokens beginning in lowercase, punctuation and comma rates, the rate of spaces before punctuation, the repeated-bigram rate, the rates of unigrams and of bigrams absent from the source article, and the digit rate.
Counts are normalized by summary length, so that no feature stands in for token count.
These cover the surface properties on which SummEval's 16 systems differ most, including the lowercasing and detokenization artifacts visible in Figure~\ref{fig:summeval_record}.

The protocol matches the one in Section~\ref{sec:probe_method}.
We standardize the features, fit \texttt{LogisticAT} at the same regularization strength on the same 1{,}280 training summaries, and score ordinal Krippendorff's $\alpha$ against the median expert rating on the same 320 test summaries.
Uncertainty comes from the paired bootstrap of Section~\ref{sec:probe_method}, so the surface probe and each judge's selective probe are scored on the same 10{,}000 resamples of the test summaries.

\begin{table}[h]
\footnotesize
\centering
\setlength{\tabcolsep}{4pt}
\begin{tabular}{lr@{\hskip 12pt}rr rr}
\toprule
& & \multicolumn{2}{c}{Gemma-3} & \multicolumn{2}{c}{Qwen-3.5} \\
\cmidrule(lr){3-4} \cmidrule(lr){5-6}
Criterion & $\alpha$ & $\Delta$ & sd$(\Delta)$ & $\Delta$ & sd$(\Delta)$ \\
\midrule
Coherence & $0.138$ & $0.543$ & $0.055$ & $0.545$ & $0.054$ \\
Consistency & $-0.058$ & $0.615$ & $0.090$ & $0.752$ & $0.078$ \\
Fluency & $-0.072$ & $0.630$ & $0.085$ & $0.672$ & $0.077$ \\
Relevance & $-0.006$ & $0.544$ & $0.048$ & $0.598$ & $0.053$ \\
\bottomrule
\end{tabular}
\caption{Ordinal agreement with the median expert rating for a probe on 13 surface properties of the summary text.
The surface-feature probe is independent of the LLM judge, so we use the same baseline agreement value in both comparisons.
Here $\Delta = \alpha_{\mathrm{Selective}} - \alpha_{\mathrm{surface}}$, using that judge's \emph{Selective} probe from Table~\ref{tab:probing_results}, so positive values favor the selective features.
sd$(\Delta)$ is the standard deviation of this paired difference over 10{,}000 bootstrap resamples of the test summaries. We score both probes on the same resample with each probe fit held fixed.}
\label{tab:surface_baseline}
\end{table}

Baseline agreement is $0.138$ on coherence and non-positive on the other three criteria.
For consistency and fluency, the probe predicts one rating level for every test summary, so its $\alpha$ reflects a constant prediction.
The selective probe has higher agreement in all eight judge-criterion pairs, with a minimum difference of $0.543$.
Every difference exceeds six paired standard deviations.
The highest agreement of the surface-feature probe is below the lowest activation-probe agreement in Table~\ref{tab:probing_results}.
The fluency result of Section~\ref{sec:probing_results} in particular does not follow from the lowercasing and detokenization artifacts that characterize this generation of summarizers.

\subsection{Selective Feature Counts in All Configurations}
\label{app:all_selected}

\begin{figure*}[ht]
\centering
\begin{subfigure}{0.9\textwidth}
    \centering
    \includegraphics[width=\textwidth]{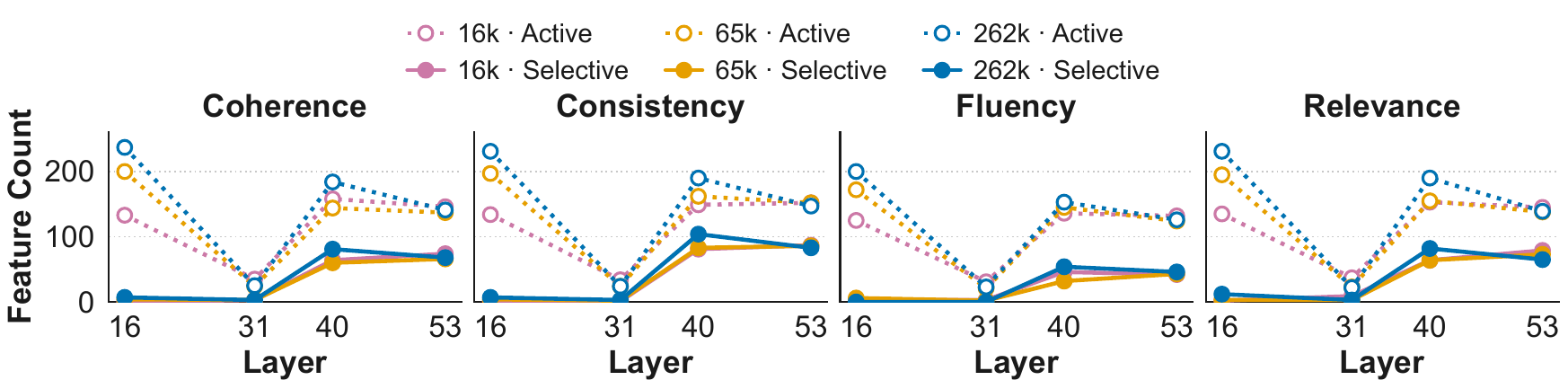}
    \caption{Gemma~3}
    \label{fig:filter_comparison_gemma}
\end{subfigure}

\begin{subfigure}{0.9\textwidth}
    \centering
    \includegraphics[width=\textwidth]{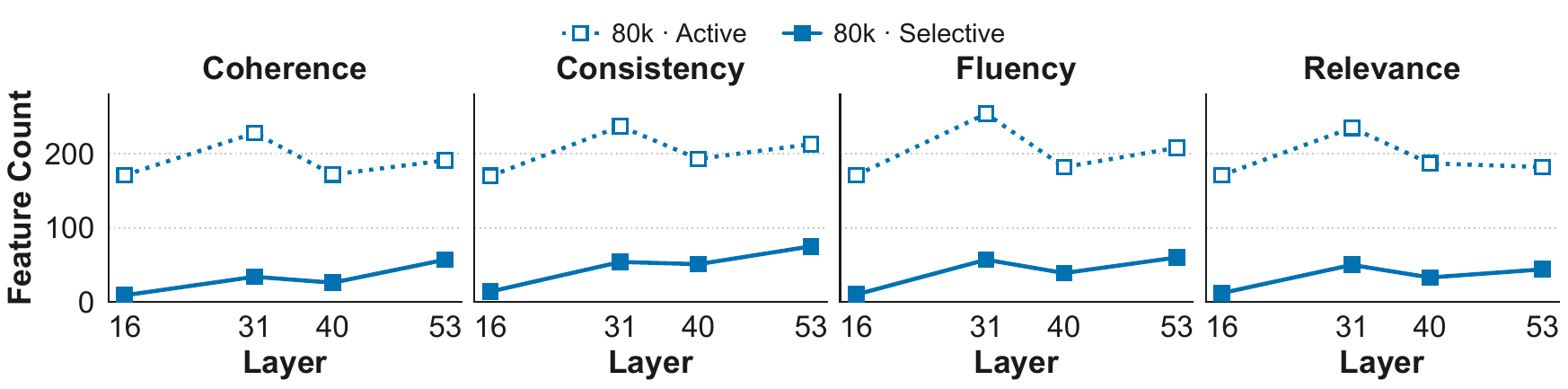}
    \caption{Qwen~3.5}
    \label{fig:filter_comparison_qwen}
\end{subfigure}
\caption{Active- and selective-feature counts by layer for each criterion, with one line per SAE width: (a) Gemma~3 (widths 16k, 65k, 262k) and (b) Qwen~3.5 (width 80k).
The main results use layer 53, at width 16k for Gemma~3 and width 80k for Qwen~3.5. In Table~\ref{tab:selective_counts} we report the selective counts at those configurations (Section~\ref{sec:feature_activation}).}
\label{fig:filter_comparison}
\end{figure*}

We compare selective-feature counts across layers and SAE widths to explain the configuration used in Section~\ref{sec:results}: layer 53 at width 16k for Gemma~3 and width 80k for Qwen~3.5.
Figure~\ref{fig:filter_comparison} gives the full grid: all four criteria, all four layers, and all three Gemma Scope~2 widths alongside the single Qwen-Scope width.
Each panel plots the number of active features (dotted lines, open points) against the number that pass the selectivity rule of Section~\ref{sec:selectivity_method} (solid lines, filled points), with one color per dictionary width.

Gemma~3 has few selective features at layers 16 and 31 for all four criteria and all three widths, and many more at layers 40 and 53.
Qwen~3.5 has higher counts at layer 31, followed by a decrease at layer 40 in every criterion.
Layer 53 has its largest counts for coherence, consistency, and fluency. For relevance, the count at layer 31 is only marginally higher.
Layer 53 is at or near the maximum for both judges on almost every criterion.
Quality-relevant representations appear to emerge late, as the residual stream accumulates task-specific information.

Selective-feature counts vary much less with width than with layer depth.
Gemma~3's counts are similar across widths except at layer 40, where the 262k SAE has more selective features.
At layer 53, the narrowest dictionary has slightly higher counts on most criteria.
The 16k SAE yields at least as many selective features as the 262k on coherence (74 against 68), consistency (87 against 83) and relevance (79 against 65), and only on fluency does it yield fewer features (42 against 46).
We therefore decided to use the smallest available width in each case, because it reduces compute and storage costs and has fewer ``dead'' features.

In Table~\ref{tab:selective_counts} we report the feature counts for the selected configurations.
Only then did we perform the probing experiments that led to our central result in Table~\ref{tab:probing_results}.
We do not report probe results at the other layers and widths, so we do not claim that our findings would apply to them.

Active- and selective-feature counts follow different patterns, so the rule is not a proxy for activation frequency.
Gemma~3's active counts are highest at layer 16 for the two wider dictionaries, precisely where almost nothing is selective.
Qwen~3.5's active-feature counts peak at layer 31 in all four criteria, before its selective counts do.

\subsection{Interpretability of Extracted SAE Features}
\label{app:neuronpedia}

Neuronpedia's published explanations for the features our rule selects have little to do with evaluation criteria, instead discussing numbers and code.

Our selection rule identifies features that distinguish high- from low-scoring summaries but does not explain what those features encode.
We therefore examined Neuronpedia's descriptions of the selected features for both judges.

We queried the Neuronpedia API\footnote{\url{https://www.neuronpedia.org}} for every selective feature at the configuration of Section~\ref{sec:results} (Gemma-3, layer 53, width $16\text{k}$).
The query covered 282 features across the four criterion-specific sets, spanning 140 distinct feature indices.
Every request succeeded.
Table~\ref{tab:neuronpedia} reports the results.

\begin{table}[ht]
\centering
\small
\begin{tabular}{lrrr}
\toprule
Criterion & $n$ & Explained & Quality \\
\midrule
Coherence   & 74  & 57 (77\%) & 9 (12\%) \\
Consistency & 87  & 69 (79\%) & 14 (16\%) \\
Fluency     & 42  & 35 (83\%) & 7 (17\%) \\
Relevance   & 79  & 59 (75\%) & 8 (10\%) \\
\midrule
Total       & 282 & 220 (78\%) & 38 (13\%) \\
\bottomrule
\end{tabular}
\caption{Neuronpedia coverage for selective Gemma-3 features at layer 53, width $16\text{k}$, for the expert-annotated selective sets used throughout Section~\ref{sec:results}.
\emph{Explained} counts features carrying at least one explanation.
\emph{Quality} counts features whose explanations contain a keyword from our list for the four SummEval criteria.
Some matches are incidental, so this count overstates how often the explanations describe summary quality.}
\label{tab:neuronpedia}
\end{table}

Of the 140 distinct features, 113 (81\%) have a Neuronpedia explanation and 27 (19\%) have none.
Table~\ref{tab:neuronpedia} counts each explained feature once for each criterion where it is selective, so the 113 features account for 220 entries.
Across those features, we found 111 distinct descriptions.
Only ``numbers and currency'' and ``three'' appear twice.
Every explanation was written by the same small language model, \texttt{gemini-2.5-flash-lite}~\citep{gemini25technicalreport}, from a feature's top-activating examples.
These descriptions are generated automatically from activation examples. They are not human annotations of summary quality in our evaluation task.

We sort the 140 distinct features by what their explanation describes.
The largest identifiable groups concern numbers or quantities (36 features, 26\%; \emph{numbers and currency}, \emph{arithmetic and sums}) and code or markup (21 features, 15\%; \emph{python code}, \emph{html structure}).
Another 6 features (4\%) describe greetings or conversational openers.
The remaining 50 explained features (36\%) concern topics with no evident relation to text evaluation.
The keyword match in Table~\ref{tab:neuronpedia} is the most favorable interpretation available, and it still reaches only 13\%.
Those 38 matches contain 22 distinct explanation strings.
Most are spurious on inspection (\emph{code structure delimiters} matches on \emph{structure}, \emph{statement of fact} on \emph{fact}).
Only a few distinct explanations relate directly to summary evaluation, such as \emph{generate coherent text}, \emph{writing style, tone, and polish}, and \emph{errors and mistakes}.

Several selective features have explanations about rating scales (\emph{ratings and scores}, \emph{rating scales 1-10 or 1-5}, \emph{numerical rating `3'}). Numerical descriptions form the largest identifiable group.
We record activations at the last token of the prompt, immediately before the judge reports a score, so we expect features tied to producing a number.
These descriptions concern score generation rather than the preceding assessment of the summary.

Neuronpedia covers our reported Gemma-3 configuration in full, but its single Qwen-3.5 source, \texttt{31-qwenscope-res-80k}, covers only layer 31. It has no entry for layer 53, which we use for the main results.
Table~\ref{tab:neuronpedia_qwen} reports the Qwen-3.5 results at layer 31.
Section~\ref{sec:feature_activation} excludes this layer from later comparisons because too few features meet the selectivity threshold.

\begin{table}[ht]
\centering
\small
\begin{tabular}{lrrr}
\toprule
Criterion & $n$ & Explained & Quality \\
\midrule
Coherence   & 34  & 31 (91\%) & 2 (6\%) \\
Consistency & 54  & 50 (93\%) & 6 (11\%) \\
Fluency     & 57  & 54 (95\%) & 6 (11\%) \\
Relevance   & 50  & 47 (94\%) & 8 (16\%) \\
\midrule
Total       & 195 & 182 (93\%) & 22 (11\%) \\
\bottomrule
\end{tabular}
\caption{Neuronpedia coverage for selective Qwen-3.5 features at layer 31, width $80\text{k}$, the only Qwen-Scope source Neuronpedia serves.
Columns are as in Table~\ref{tab:neuronpedia}, again for the expert-annotated selective sets.}
\label{tab:neuronpedia_qwen}
\end{table}

Qwen-3.5 has greater explanation coverage than Gemma-3, at 93\% versus 78\%, but a similarly small share matches the criterion keywords: 11\% versus 13\%.
Some explanations are even less related to the task (\emph{C\# System namespaces}, \emph{Star DC Comics}).
Numerical descriptions account for only 9\% of Qwen-3.5 features, compared with the much larger group in Gemma-3 at layer 53.
That is consistent with interpreting the numeric features as tied to reporting a score, since layer 31 sits further from the output.
This comparison changes both model and layer, so their effects cannot be separated.

Section~\ref{sec:probing_results} shows that the selective features help predict expert ratings.
An explanation generated from a feature's top-activating examples on a general web corpus describes the feature's behavior on that corpus, not its role in a specific downstream task.
A feature associated with numerals in arithmetic contexts may also help predict summary quality during the judging task.
Explaining that role requires methods specific to summary evaluation~\citep{paulo2024automatically}, which we leave to future work.

\subsection{Inter-Criterion Selective Feature Overlap vs.\ Feature Index}
\label{app:overlap_vs_feat_id_all_pairs}

Gemma~3's SAEs were trained with a Matryoshka objective, under which lower-index features should be more "universal" (shared across criteria) and higher-index features more fine-grained and criterion-specific.
Qwen~3.5's SAEs were not, so they should show no such trend.
We test whether criteria share fewer features at higher indices, as Gemma~3's feature ordering would predict.
We divide each dictionary into five equal bins by feature index.
Within each bin, we compute Jaccard overlap for each pair of criteria, first for active features and then for the selective features identified in Section~\ref{sec:exp2}.
Gemma Scope~2 does not report the boundaries between nesting levels, so a bin may span more than one level.

\begin{figure}[ht]
\centering
\includegraphics[width=0.92\linewidth]{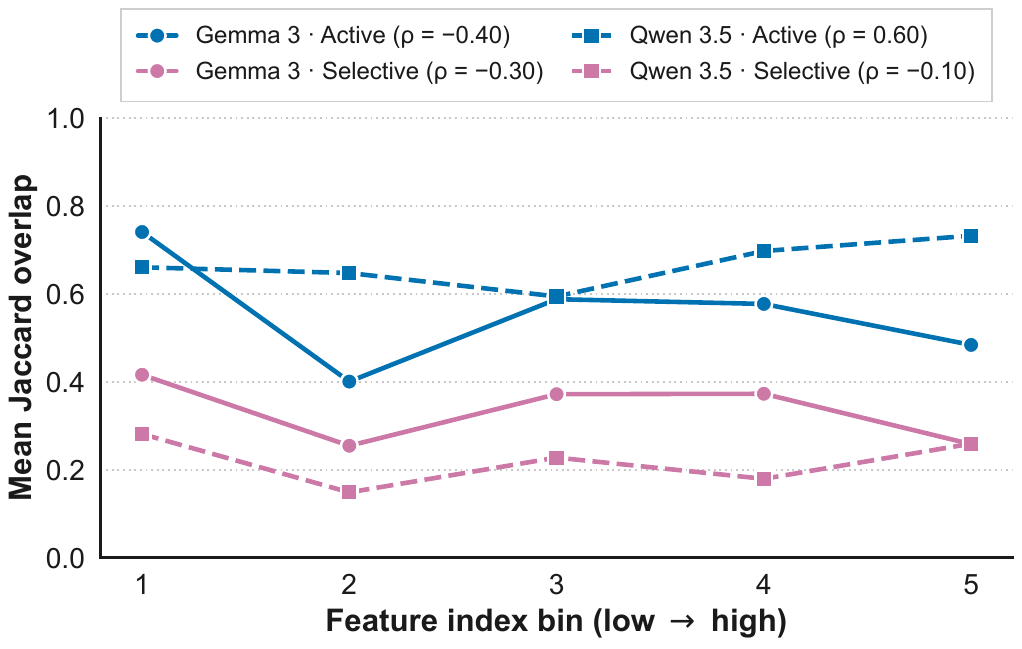}
\caption{Mean Jaccard overlap across the six pairs of criteria, from the lowest feature indices (bin 1) to the highest (bin 5).
For Gemma~3, active-feature overlap is highest in bin 1 and lowest in bin 2.
For Qwen~3.5, overlap generally rises with feature index for active features and shows no clear trend for selective features.}
\label{fig:matryoshka}
\end{figure}

Averaged over the six pairs, Gemma~3's active-feature Jaccard peaks in the first bin and drops in later bins, though not monotonically (Figure~\ref{fig:matryoshka}).
The second bin is the lowest, after which similarity rises before falling again.
Qwen~3.5's active-feature overlap increases with feature index ($\rho = 0.60$), while its selective-feature overlap shows no clear trend ($\rho = -0.10$).

With only five bins, a Spearman correlation of $0.60$ in either direction is not statistically significant.
The observed trends therefore offer little evidence for the Matryoshka hypothesis.

\begin{figure*}[ht]
\centering
\begin{subfigure}{0.32\textwidth}
\includegraphics[width=\textwidth]{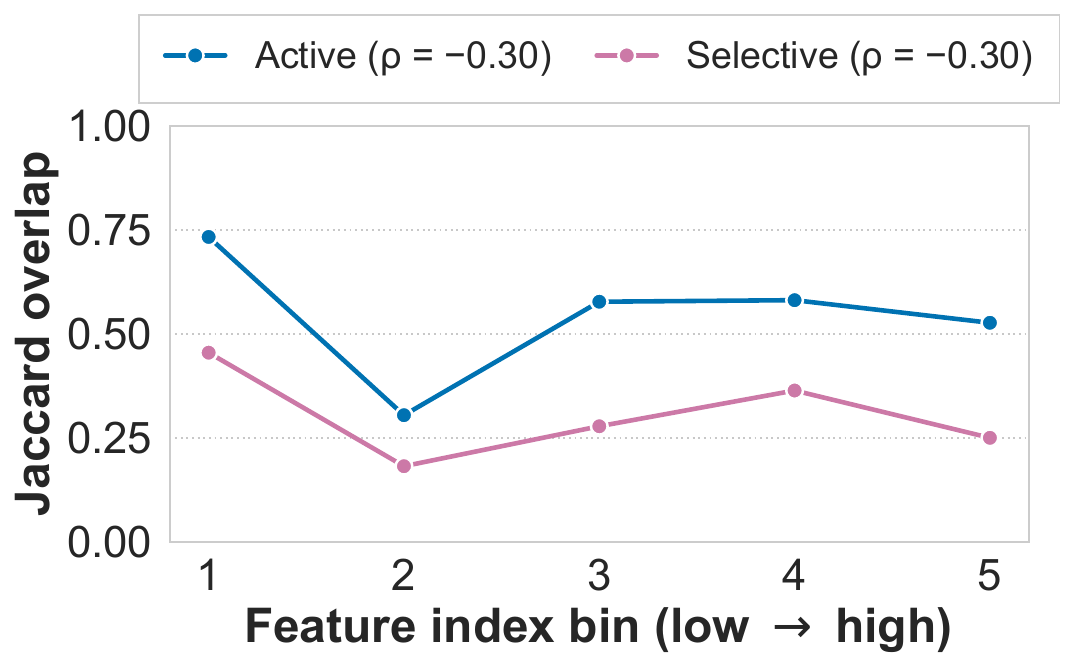}
\caption{Coherence vs Consistency}
\label{fig:jaccard_cc}
\end{subfigure}
\hfill
\begin{subfigure}{0.32\textwidth}
\includegraphics[width=\textwidth]{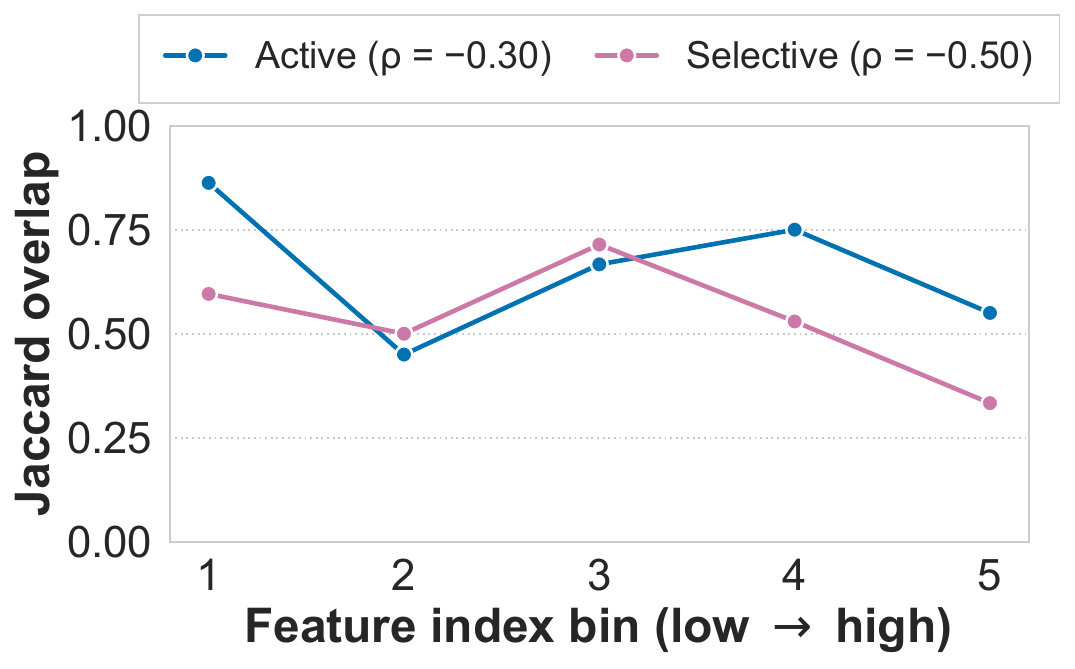}
\caption{Coherence vs Relevance}
\label{fig:jaccard_cr_gemma}
\end{subfigure}
\hfill
\begin{subfigure}{0.32\textwidth}
\includegraphics[width=\textwidth]{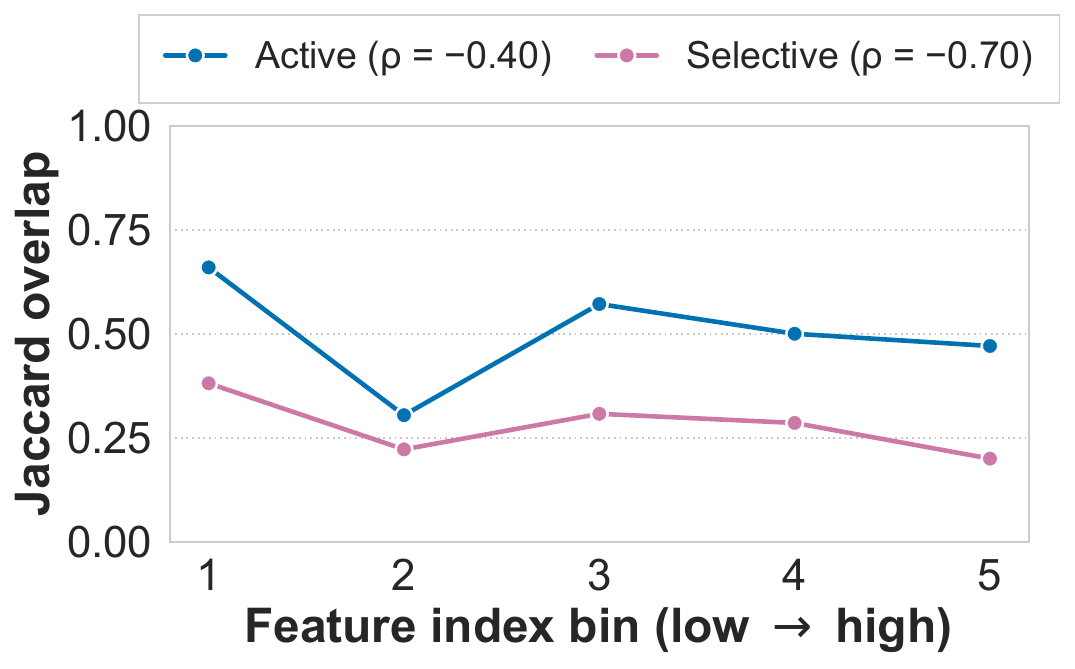}
\caption{Coherence vs Fluency}
\label{fig:jaccard_cf_gemma}
\end{subfigure}

\medskip
\begin{subfigure}{0.32\textwidth}
\includegraphics[width=\textwidth]{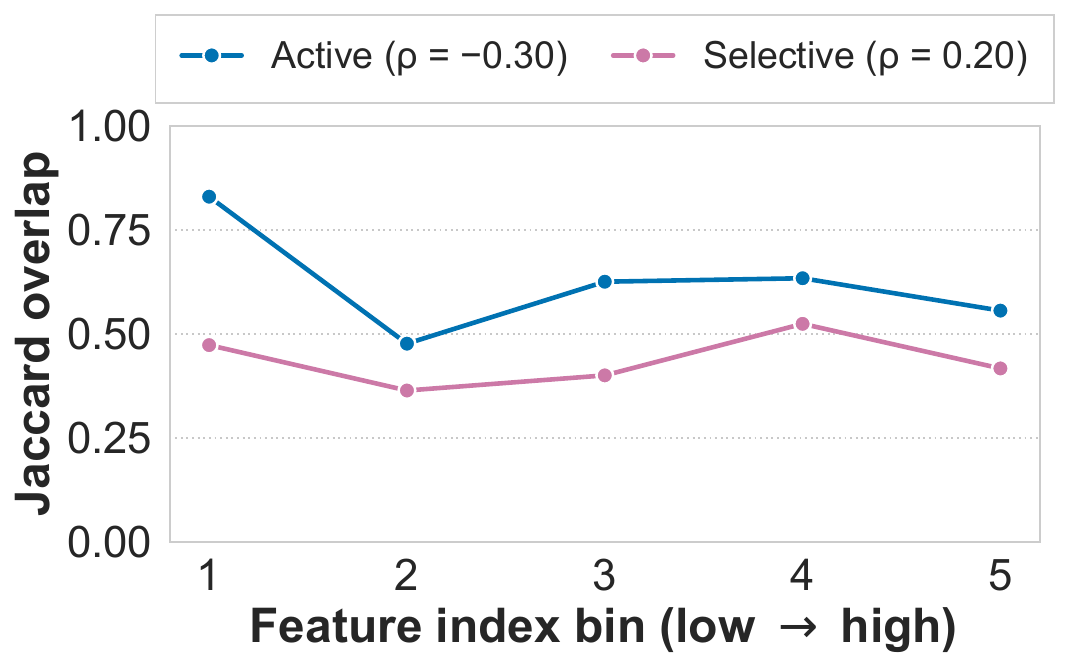}
\caption{Consistency vs Relevance}
\label{fig:jaccard_cr2_gemma}
\end{subfigure}
\hfill
\begin{subfigure}{0.32\textwidth}
\includegraphics[width=\textwidth]{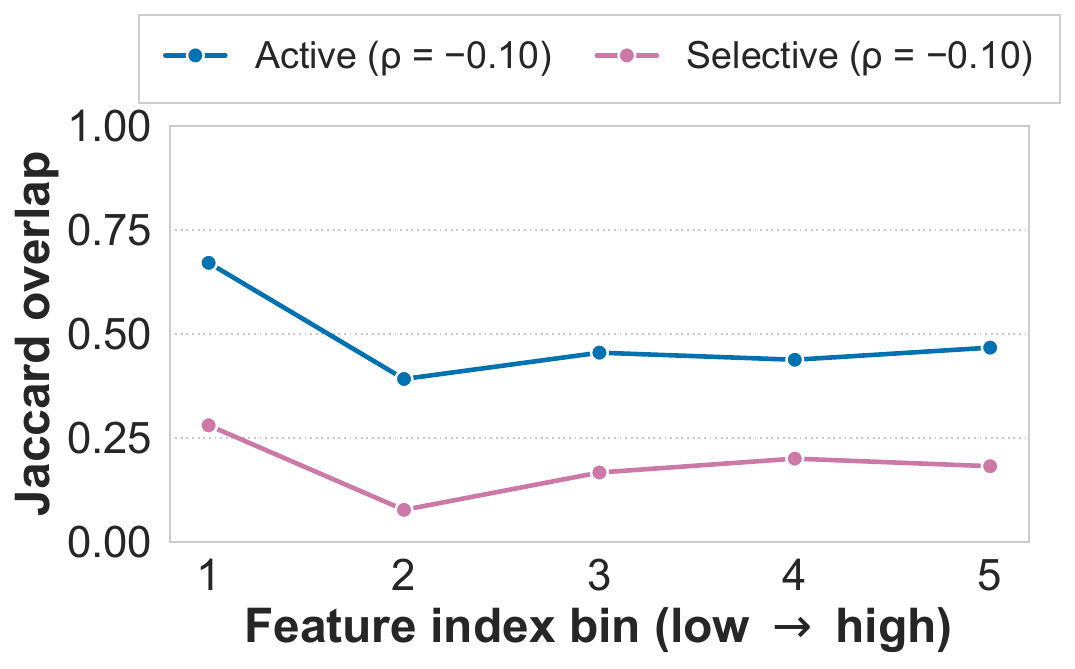}
\caption{Consistency vs Fluency}
\label{fig:jaccard_cf2_gemma}
\end{subfigure}
\hfill
\begin{subfigure}{0.32\textwidth}
\includegraphics[width=\textwidth]{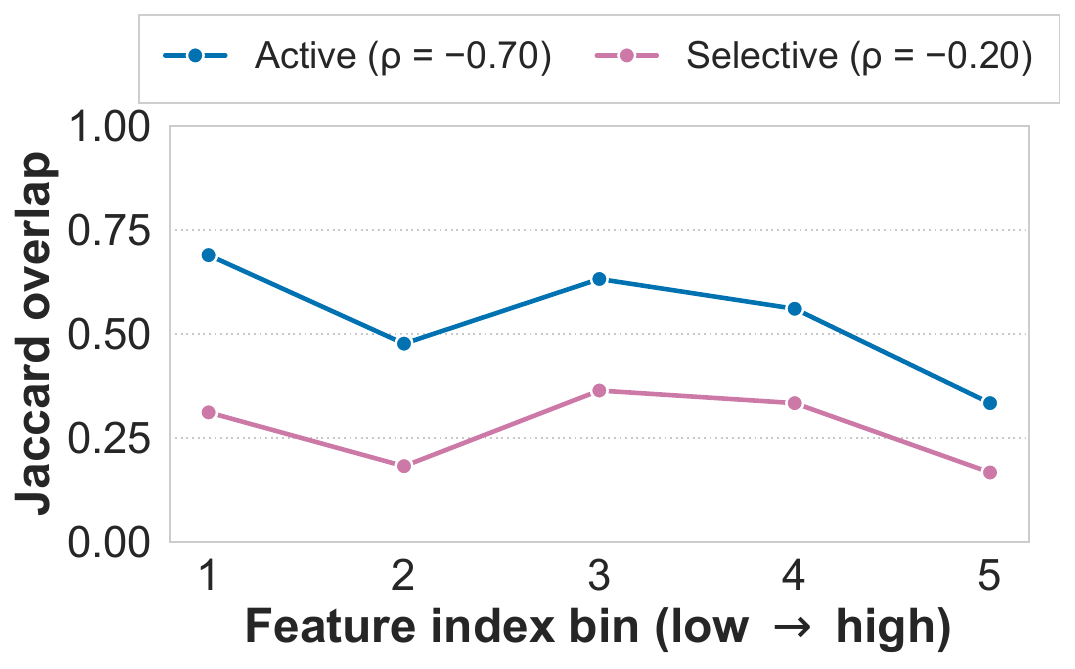}
\caption{Relevance vs Fluency}
\label{fig:jaccard_rf_gemma}
\end{subfigure}
\caption{Each panel shows Jaccard overlap for one pair of criteria in Gemma~3.
The two curves show overlap for active features and for selective features.
We group the 16{,}384 feature indices into five consecutive bins of about 3{,}277 indices each.}
\label{fig:jaccard_pairs_gemma}
\end{figure*}

\begin{figure*}[ht]
\centering
\begin{subfigure}{0.32\textwidth}
\includegraphics[width=\textwidth]{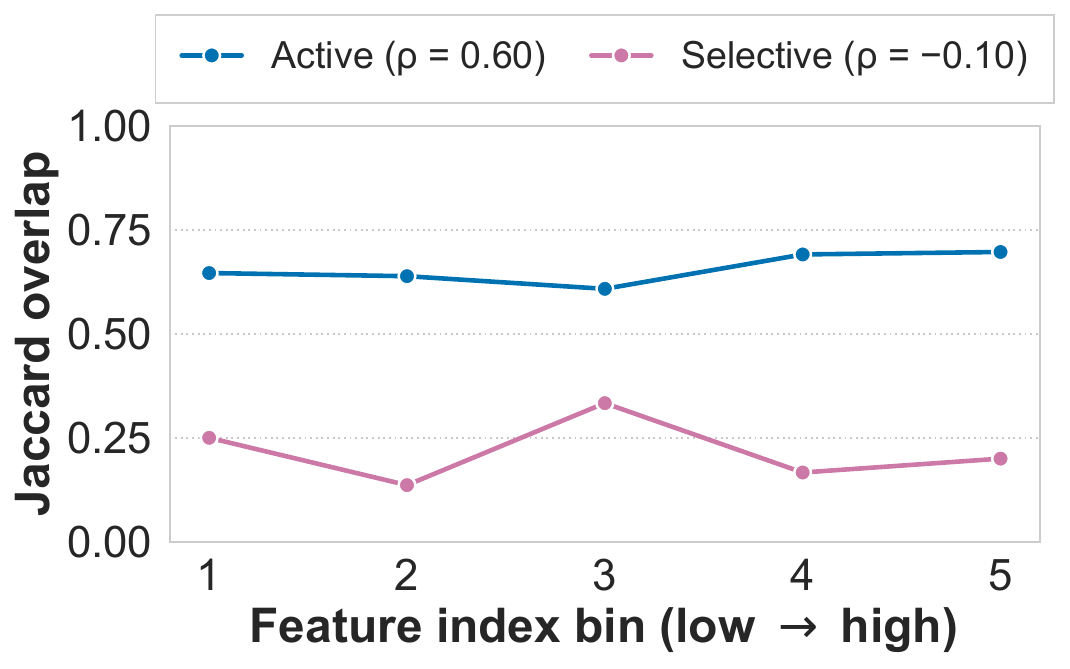}
\caption{Coherence vs Consistency}
\label{fig:jaccard_cc_qwen}
\end{subfigure}
\hfill
\begin{subfigure}{0.32\textwidth}
\includegraphics[width=\textwidth]{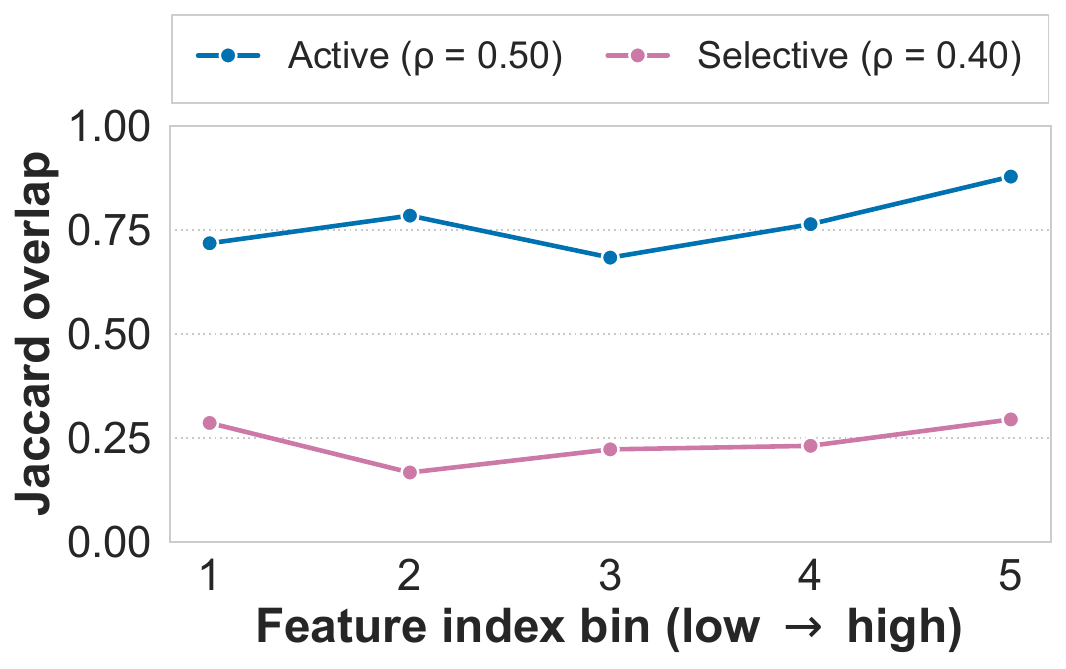}
\caption{Coherence vs Relevance}
\label{fig:jaccard_cr_qwen}
\end{subfigure}
\hfill
\begin{subfigure}{0.32\textwidth}
\includegraphics[width=\textwidth]{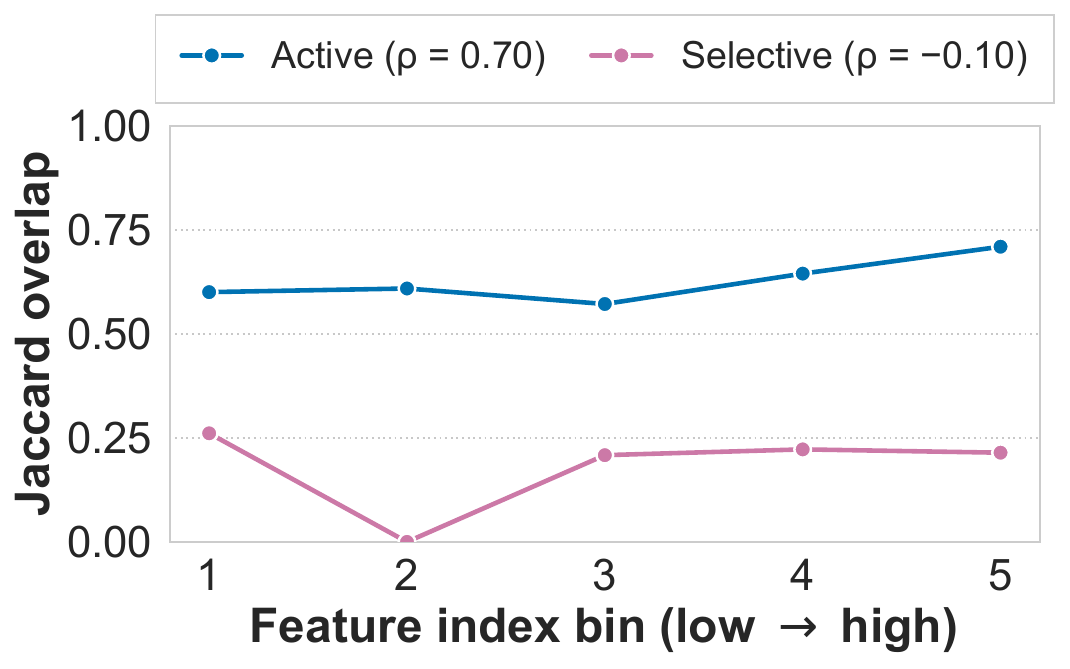}
\caption{Coherence vs Fluency}
\label{fig:jaccard_cf_qwen}
\end{subfigure}

\medskip
\begin{subfigure}{0.32\textwidth}
\includegraphics[width=\textwidth]{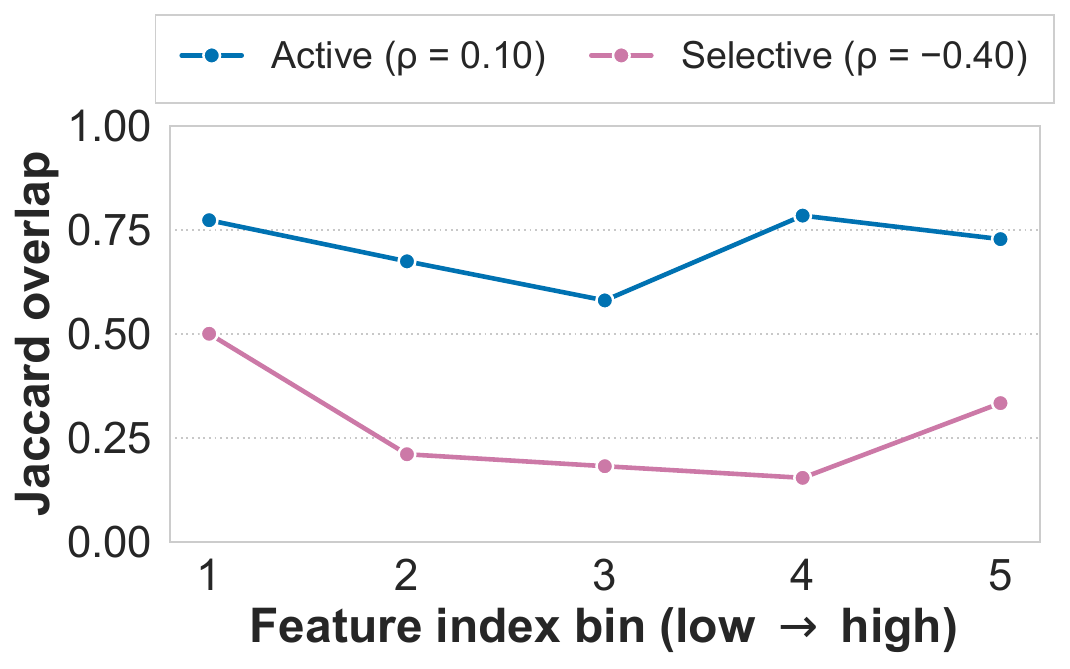}
\caption{Consistency vs Relevance}
\label{fig:jaccard_cr2_qwen}
\end{subfigure}
\hfill
\begin{subfigure}{0.32\textwidth}
\includegraphics[width=\textwidth]{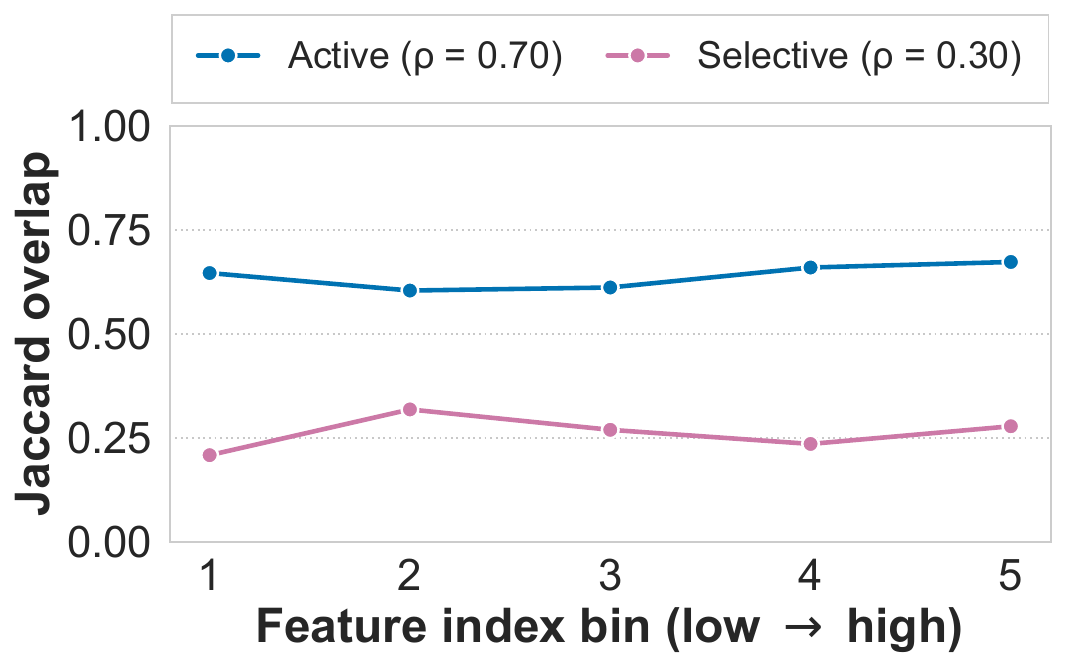}
\caption{Consistency vs Fluency}
\label{fig:jaccard_cf2_qwen}
\end{subfigure}
\hfill
\begin{subfigure}{0.32\textwidth}
\includegraphics[width=\textwidth]{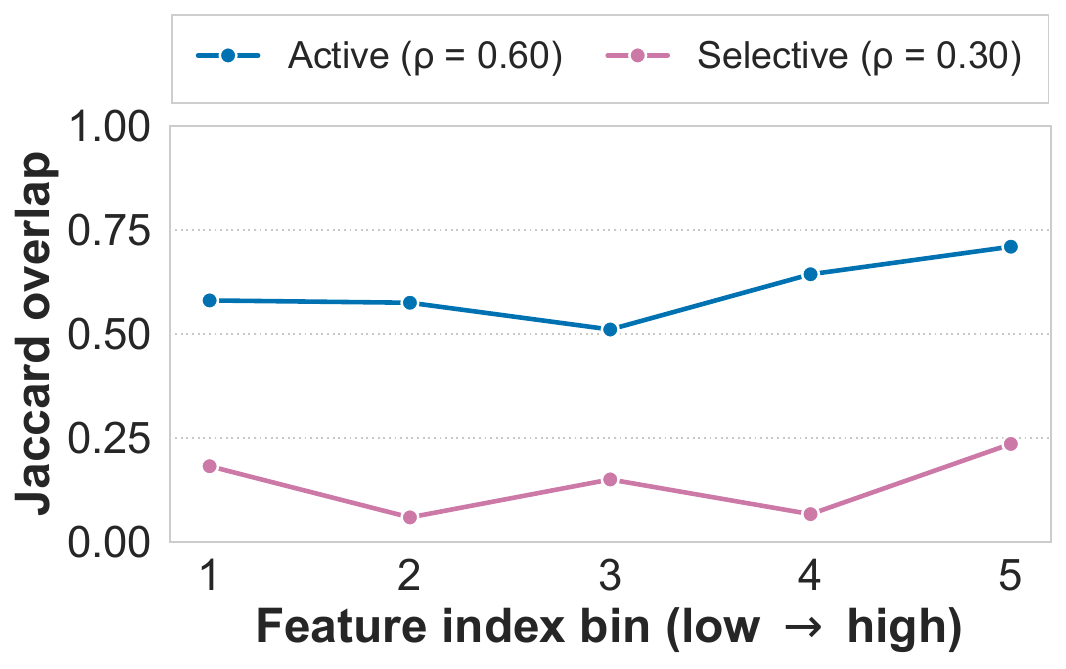}
\caption{Relevance vs Fluency}
\label{fig:jaccard_rf_qwen}
\end{subfigure}
\caption{Jaccard overlap by feature index for each criterion pair, shown separately for active and selective features in Qwen~3.5, with 5 bins of 16,384 features each.}
\label{fig:jaccard_pairs_qwen}
\end{figure*}

We also examine each of the six criterion pairs to determine whether the aggregate pattern occurs across pairs, for Gemma~3 (Figure~\ref{fig:jaccard_pairs_gemma}) and Qwen~3.5 (Figure~\ref{fig:jaccard_pairs_qwen}).

For Gemma~3, most criterion pairs follow the aggregate pattern.
Overlap is highest in the first (coarsest) feature bin and generally declines toward the later bins, though not monotonically.
The second-bin dip in the mean curve is visible in most of the six pairs, so it is a broadly shared feature of the dictionary's structure rather than an artifact of one outlier pair.
The last bin's overlap is similar to, and sometimes lower than, the second bin's. The curve does not have a clear minimum at the highest feature indices, as a strict Matryoshka hierarchy would predict.
Pairs involving relevance are noisier, since relevance has fewer selective features and its bin-level Jaccard estimates are more sensitive to individual membership.
Every active-feature curve declines ($\rho$ from $-0.10$ to $-0.70$), and five of the six selective curves are downward or flat, with consistency-relevance the one weak exception ($\rho = 0.20$).

Qwen~3.5 shows a different pattern.
Its selective-feature curves have no shared direction ($\rho$ from $-0.40$ to $+0.40$), and active-feature overlap increases with index in five of the six pairs ($\rho \geq 0.50$).
No pair reproduces Gemma~3's first-bin peak or its early dip.
The dictionaries show different overlap patterns, but this comparison does not identify the cause.

